\documentclass[10pt,conference]{IEEEtran}
\usepackage{cite}
\usepackage{amsmath,amssymb,amsfonts}
\usepackage{algorithmic}
\usepackage{algorithm}
\usepackage{graphicx}
\usepackage{textcomp}
\usepackage[table]{xcolor}
\usepackage[hyphens]{url}
\usepackage{fancyhdr}
\usepackage{enumitem}
\usepackage{comment}
\usepackage{array}
\usepackage{booktabs}
\usepackage{listings}
\usepackage{hyperref}
\usepackage{cleveref}

\newcommand{\system}{AtumAI}
\newcommand{\compiler}{Datacenter Task Compiler}
\newcommand{\loopAgent}{Evolutionary Design Discovery Loop}

\definecolor{irkey}{HTML}{1F6FB2}
\definecolor{irtype}{HTML}{0B7A75}
\definecolor{ircomment}{HTML}{8A8A8A}
\definecolor{irbg}{HTML}{F7F9FC}
\definecolor{irframe}{HTML}{D5DEE9}
\lstdefinestyle{ir}{
  basicstyle=\ttfamily\footnotesize,
  backgroundcolor=\color{irbg},
  frame=single,
  framerule=0.4pt,
  rulecolor=\color{irframe},
  keywordstyle=\color{irkey}\bfseries,
  keywordstyle=[2]\color{irtype},
  commentstyle=\color{ircomment}\itshape,
  morecomment=[l]{\#},
  morekeywords={task_type,decision_variables,objectives,constraints,execution_budget,evaluation,characterization},
  morekeywords=[2]{int,float,minimize,loose,full,hard},
  columns=fullflexible,
  keepspaces=true,
  showstringspaces=false,
  breaklines=true,
  xleftmargin=6pt,
  aboveskip=2pt,
  belowskip=2pt,
  literate={<=}{{$\le$}}1 {>=}{{$\ge$}}1,
}

\newcommand*{\rom}[1]{\uppercase\expandafter{\romannumeral #1\relax}}

\title{\emph{\system{}}: A Principled Framework for Agentic Generation of Datacenter Control-Plane Policies}

\newcommand\paperauthors{Qiushi Lin$^{\dagger}$, Chaojie Zhang$^{*}$, \'I\~nigo Goiri$^{*}$, Aditya Akella$^{\dagger}$, Ricardo Bianchini$^{*}$, Jovan Stojkovic$^{\dagger}$}
\newcommand\paperaffiliation{$^{\dagger}$The University of Texas at Austin \qquad $^{*}$Microsoft Azure}

\author{
  \IEEEauthorblockN{\paperauthors{}}
  \IEEEauthorblockA{\paperaffiliation{}}
}

\begin{document}
\maketitle

\thispagestyle{plain}
\pagestyle{plain}

\begin{abstract}
The efficiency of a datacenter rests on its control plane, i.e., the policies that decide how to make the best use of
the available hardware.
Designing these policies, however, is increasingly hard: the hardware-software stack grows faster than the
engineers who run it can reason about, the design space is so vast and interdependent that experts miss most of the
opportunities it holds, and prototyping even a single candidate policy takes months.
Agentic AI promises to automate this search.
Off the shelf, however, it falls short on three fronts.
It is not \emph{formal}: with no structured, searchable statement of the problem, the search has little
structure to exploit and hard constraints are not guaranteed.
It is not \emph{transferable}: each task is solved from scratch, so nothing learned on one task carries to the next.
Finally, it is not \emph{systematic}: relying on the LLM as the sole source of candidates, it explores only a narrow, biased slice
of the design space and settles into local optima.

We introduce \system{}, a framework that generates datacenter control-plane policies with agentic AI, making the process
formal, transferable, and systematic where off-the-shelf systems are not.
From a goal stated in plain language, \system{} autonomously proposes, tests, and refines candidate policies until
one satisfies the request.
It does so through two components.
The \emph{\compiler{}} automates problem formulation: it compiles the request into a formal, machine-checkable, and
searchable specification of the task's objectives, constraints, decision variables, and evaluation
methodology, grounded in the target's mined workload and platform characterization.
The \emph{\loopAgent{}} then searches this specification, expanding the search beyond the LLM itself: a diffusion
model explores structurally distinct designs, an evolutionary algorithm tunes their parameters, and a surrogate
model filters candidates before costly evaluation.
Together, they reduce onboarding a new task from months of engineering to writing its description. 

We evaluate \system{} on three control-plane tasks with distinct problem scopes, design spaces, and trade-offs:
\emph{workload placement}, \emph{resource scaling}, and \emph{power management}.
Across all tasks, the policies generated by \system{} consistently outperform
expert-engineered baselines, showing that a single \system{} pipeline generalizes across control-plane optimization problems.
\end{abstract}

\definecolor{tabhead}{HTML}{33415C}
\definecolor{tabrow}{HTML}{EEF2F8}
\definecolor{tabaccent}{HTML}{1F6FB2}

\begin{table*}[t]
\centering
\footnotesize
\setlength{\tabcolsep}{10pt}
\renewcommand{\arraystretch}{1.4}
\caption{Where off-the-shelf agentic systems fall short on control-plane policy design, and how \system{} answers each.}
\vspace{-3mm}
\label{tab:novelty}
\begin{tabular}{@{}>{\raggedright\arraybackslash}p{0.13\linewidth} >{\raggedright\arraybackslash}p{0.37\linewidth} >{\raggedright\arraybackslash}p{0.36\linewidth}@{}}
\toprule
\rowcolor{tabhead}
\textcolor{white}{\textbf{Missing ingredient}} & \textcolor{white}{\textbf{Resulting issue in prior systems}} &
\textcolor{white}{\textbf{\system{}'s answer}} \\
\midrule
\textcolor{tabaccent}{\textbf{Formal}} & No structured, searchable problem grounded in the target, so hard constraints are not
guaranteed. & The \compiler{} compiles the request into a standardized, searchable, machine-checkable IR grounded in the target's real workload and platform. \\
\rowcolor{tabrow}
\textcolor{tabaccent}{\textbf{Transferable}} & Each task is bespoke, so nothing learned on one task carries to the next. & The shared IR and library let knowledge distilled on one task transfer to the next. \\
\textcolor{tabaccent}{\textbf{Systematic}} & The LLM is the only explorer, so the search is narrow, biased, and trapped in
local optima. & The \loopAgent{} expands beyond the LLM with diffusion (conceptual) and EA (numeric) search, plus a surrogate filter. \\
\bottomrule
\end{tabular}
\vspace{-3mm}
\end{table*}

\section{Introduction}
\label{sec:intro}

Datacenters are the backbone of modern computing, powering microservices,
web search, storage, analytics, AI training and inference~\cite{warehouseScaleComp,accelerometer,tapas,smartoclock,googlewebsearch,deathstarbench,netflix,micromanycore,metaMicro}.
Their efficiency, cost, and reliability rest on the \emph{control plane}:
the software policies that decide how efficiently the underlying hardware is used~\cite{dynamoFB,thunderbolt,protean,sinanML,distributedcaching},
either by extracting more useful work from hardware
(e.g., workload placement, resource scaling, caching, and traffic routing) or
by acting on the hardware directly
(e.g., power management).
Each such policy governs a vast design space with competing objectives (e.g., server utilization, tail latency, cost, and energy) under hard constraints
(e.g., service-level objectives, capacity limits, and hardware and infrastructure requirements such as power envelopes).
As hardware-software stack scales and diversifies~\cite{microsoftchip,googlechip,amazonchip} while workloads keep evolving~\cite{dcperf,mahar2023workloadbehaviordrivenmemory,microsoftworkloads}, this space expands combinatorially.
Thus, the traditional, expert-driven way of designing and tuning control-plane policies no longer keeps
pace~\cite{resourcecentral,cemri2026adaevolveadaptivellmdriven}.
This causes a steady stream of missed efficiency opportunities that, at datacenter scale, waste
millions of dollars and megawatts.

Recent advances in agentic AI offer a path forward.
\emph{AI agents} are LLM-powered systems that can reason about goals, interact with their environment, and autonomously execute multi-step actions~\cite{react,schick2023toolformerlanguagemodelsteach}.
They can generate and run code, invoke external tools, and coordinate complex workflows.
These capabilities are already transforming software engineering~\cite{sweagent,metagpt,kernelevolve} and accelerating scientific discovery~\cite{novikov2025alphaevolvecodingagentscientific}.
We argue that they can similarly transform datacenter control-plane policy design.
Rather than replacing human experts, agents can amplify their creativity, exploring orders of magnitude more
of the design space than a human can, and freeing engineers to focus on framing problems and judging results.

However, simply pointing an agent at a control-plane problem is not enough.
Off-the-shelf agentic systems, and even the most capable LLM-driven evolutionary-search
frameworks~\cite{novikov2025alphaevolvecodingagentscientific,cemri2026adaevolveadaptivellmdriven,skydiscover},
fall short on three fronts (\Cref{tab:novelty}).
First, they are not \emph{formal}:
the problem is encoded implicitly in hand-written code and fitness functions, and not in a structured, searchable specification of objectives, constraints, and decision variables grounded in the target workload and platform. 
Hence, the search has little structure to exploit, hard constraints are not guaranteed, and policies that perform well in the optimization loop can fail in production. 
Second, they are not \emph{transferable}:
each task requires bespoke scaffolding and is solved largely from scratch, limiting the reuse of knowledge, abstractions, and search strategies across domains. 
Third, they are not \emph{systematic}:
candidate solutions are generated by the LLM, causing the search to inherit model biases, explore only a narrow region of the design space, and converge prematurely to local optima.
Moreover, the performance is tightly coupled to the capabilities of a particular model. 
These limitations reinforce one another:  without a \emph{formal} problem representation, there is little structure for a \emph{systematic} search to exploit.  

\vspace{2pt}
\noindent \textbf{Our work.}
We present \emph{\system{}}, a principled framework for agentic generation of datacenter control-plane policies.
An engineer describes a policy problem in natural language, and \system{} autonomously generates, evaluates,
validates, and iteratively refines candidate policies until one satisfies the specified goals.
Because the process is fully automated and inexpensive, \system{} is not limited to one-time deployment.
It can continuously re-run as workloads, hardware, and operating conditions evolve, enabling a \emph{self-evolving datacenter}.

\system{} rests on two ideas realized by two dedicated components. 
The first idea is to treat problem \emph{formulation} as a compilation task.
\system{} realizes it through the \emph{\compiler{}}.
The compiler makes the problem \emph{formal} and \emph{transferable}.
It provides formalism by compiling the request into a standardized, searchable, and machine-checkable intermediate representation (IR) of the objectives, constraints, decision variables, and evaluation methodology, grounded in the target's mined workload and platform characterization.
It enables transfer through a shared library: expertise distilled on one problem carries to the next, so onboarding a new task extends the library rather than starting over. 

The second idea is to \emph{search} a design space that reaches beyond the LLM generated policies.
\system{} realizes it through the \emph{\loopAgent{}}.
The loop makes the search \emph{systematic}.
It broadens the search with a diffusion model for structural exploration and an evolutionary algorithm for parameter tuning.
As the search space is now significantly larger, \system{} prunes it via a surrogate model, discarding unpromising candidates before costly evaluation. 

We demonstrate \system{} on three use cases with distinct trade-offs.
For \emph{workload placement}, which balances server utilization, placement success rate, workload performance, and scheduling throughput~\cite{borg,protean,resourcecentral}, \system{} improves placement success rate by 17\% and scheduler throughput by 8\% over the expert-engineered baseline.
For \emph{resource scaling}, which trades allocated cores and memory against SLO
violations~\cite{luo2022Prediction,autopilot,sinanML}, \system{} improves cost efficiency by 24\% while keeping SLO violations at 1.3\%.
For \emph{power management}, which trades provisioned power against throughput under a per-service accuracy
floor~\cite{dynamoFB,thunderbolt,smartoclock}, \system{} cuts power by 21\% while improving throughput
by 17\%.

Across all three, \system{} produces high-quality policies from the \emph{same} framework, exceeding
expert-engineered baselines while compressing design and testing effort from months to hours. Importantly, domain knowledge is encoded once
and transferred across diverse control-plane problems without redesigning the policy-generation process.

\vspace{2pt} \noindent \textbf{Summary.}
This paper makes the following contributions:
\begin{itemize}[leftmargin=*,topsep=2pt,itemsep=1pt]
    \item \system{}, a principled framework that generates datacenter control-plane policies by compiling an informal task description into a specialized agentic workflow.
    \item \compiler{}, which lowers a natural-language task description into a searchable, machine-checkable IR of objectives, constraints, decision variables, and evaluation methodology, grounded in real workload and platform.
    \item \loopAgent{}, a search that expands the design space beyond LLM proposals with a diffusion model for structural exploration and an evolutionary algorithm for parameter tuning, and a surrogate model for fast filtering before high-fidelity evaluation.
    \item A demonstration of \system{} on three use cases: workload placement, resource scaling, and power management.
\end{itemize} 
\section{Background and Motivation}
\label{sec:background}

\subsection{The Datacenter Control Plane}
\label{sec:bg-controlplane}

The \emph{control plane} is the software that turns datacenter resources (servers, CPU cores, memory, and power) into useful work.
Its goal is to maximize the value extracted from the underlying hardware (e.g., more work per server, watt, and dollar) while respecting platform constraints.
It consists of a collection of \emph{policies}, each controlling a different mechanism (e.g., workload placement, resource allocation, or fleet-wide power management).

\vspace{2pt} \noindent \textbf{Example policies.}
Despite targeting different mechanisms, these policies share a common formulation:
they control a set of \emph{decision variables}, optimize one or more \emph{objectives}, satisfy \emph{constraints}, and are evaluated on representative workloads.
This shared structure suggests the possibility of a general framework for policy generation.
We use three representative policies as running examples throughout the paper.

\vspace{2pt} \noindent \emph{Workload placement.}
It decides which server each workload runs on.
Cluster managers such as Borg~\cite{borg}, Protean~\cite{protean}, and their successors~\cite{resourcecentral}
pack workloads onto machines while trading off server utilization, placement success rate, the performance of
workloads once scheduled, and the scheduler's own throughput, all under capacity and colocation constraints.
Decades of engineering have produced sophisticated heuristics and predictive models to navigate these tradeoffs.

\vspace{2pt} \noindent \emph{Resource scaling.}
It decides how many resources (e.g., CPU cores and memory) each service receives over time.
Systems such as Autopilot~\cite{autopilot}, and prediction- and ML-driven
schemes~\cite{luo2022Prediction,sinanML} continuously right-size allocations, trading the resources granted against
the risk of SLO violations when a service is starved.
These systems combine forecasting, feedback control, and hand-tuned safety margins.

\vspace{2pt} \noindent \emph{Power management.}
It decides how aggressively workloads may consume the power provisioned for a cluster.
Fleet-wide power managers such as Dynamo~\cite{dynamoFB}, Thunderbolt~\cite{thunderbolt}, SmartOClock~\cite{smartoclock}, and TAPAS~\cite{tapas}
oversubscribe power to raise utilization while
bounding the frequency of power-capping events, under strict power envelopes.

\vspace{2pt} \noindent \textbf{Limits of manual policy engineering.}
These systems are the product of substantial expert effort and are deployed across datacenter fleets.
However, the way they are built no longer keeps pace.
Each policy governs a combinatorial, interdependent design space that, as the hardware-software stack grows in scale and heterogeneity, expands faster than experts can explore.
Every candidate takes months to prototype, test, and evaluate.
Worse, a hand-tuned policy is frozen the moment it ships and slowly goes stale as workloads shift and hardware is refreshed, causing efficiency to erode over time. 

A more scalable approach is needed.
Rather than manually designing and tuning each policy, we seek to automatically generate policies from high-level goals.
Realizing this vision requires two capabilities.
First, the system must translate an engineer's intent (e.g., ``reduce resource usage without increasing latency'') into a precise policy specification grounded in the target platform.
Second, it must search the resulting design space broadly enough to discover high-quality solutions rather than settle for local optima.
Recent advances in \emph{agentic AI} make both capabilities practical.

\subsection{Agentic Systems for Design and Discovery}
\label{sec:bg-agentic}

An \emph{AI agent} is an LLM-powered system that perceives its environment, reasons about goals, and takes
multi-step actions (e.g., generating and executing code or invoking external
tools) to achieve them~\cite{react,schick2023toolformerlanguagemodelsteach,agenticAI}.
These capabilities have begun to transform how software and systems are built.

\vspace{2pt} \noindent \textbf{Learning for systems and architecture.}
Before agentic AI, machine learning was applied to individual control-plane and architecture mechanisms (e.g., resource scaling~\cite{sinanML}, branch prediction~\cite{perceptron}, prefetching~\cite{prefetcher,pmlr-v80-hashemi18a}, cache
replacement~\cite{replacement}, device placement~\cite{rl-device} and chip
floor planning~\cite{mirhoseini2021graph}).
These efforts show that automated search can rival hand-tuned heuristics, but each targets a single mechanism with a bespoke model and pipeline that transfers to no other.

\vspace{2pt} \noindent \textbf{LLMs and agentic AI for code and systems.}
LLMs have been applied to code generation~\cite{chen2021evaluatinglargelanguagemodels,starcoder,deepseek-coder}, program repair~\cite{xia2023apr}, compiler optimization~\cite{cummins2024}, database tuning~\cite{gptuner}, debugging~\cite{Levin_2025}, and hardware description generation~\cite{thakur2023verigen}.
Multi-agent frameworks such as SWE-agent~\cite{sweagent} and MetaGPT~\cite{metagpt} coordinate multiple agents to solve software engineering tasks, typically optimizing for functional correctness through test-driven feedback loops~\cite{llmcodegen,liu2023codegeneratedchatgptreally}.
ECO-LLM~\cite{lin2025ecollmdrivenefficientcode} extends this paradigm to performance, refactoring code based on historical anti-patterns.
However, these systems focus on correctness or narrowly defined performance objectives and operate in open-loop or tightly scoped optimization settings.

\vspace{2pt} \noindent \textbf{Agentic kernel generation and optimization.}
A growing body of work uses LLM agents to generate and optimize compute kernels under measured performance feedback.
KernelEvolve~\cite{kernelevolve}, the AI CUDA Engineer~\cite{aicudaengineer}, GEAK~\cite{geak}, and
Kevin~\cite{kevin} iteratively edit kernels, using runtime or profiler signals as a reward to converge toward faster
implementations.
This shows the power of closed-loop agentic optimization, but its target is narrow: a single objective optimized
against a fixed reward.
Control-plane policies instead demand multiple competing objectives under hard constraints that are not given up
front, but must be extracted from a plain-language goal and grounded in the platform before any search can begin.

\vspace{2pt} \noindent \textbf{LLM-driven evolutionary search.}
The next step pairs LLMs with evolutionary search, evolving programs against a fitness
function.
AlphaEvolve~\cite{novikov2025alphaevolvecodingagentscientific} and AdaEvolve~\cite{cemri2026adaevolveadaptivellmdriven}
mutate a human-written program and select variants by score, and SkyDiscover~\cite{skydiscover} makes this loop
reusable via modular components and a programmable control interface.
Two limitations, however, keep these systems from serving general control-plane design.
First, the search reaches only as far as the LLM proposes: every candidate originates from the model, so the loop
inherits its biases and fixates on a narrow region of the design space instead of exploring diverse policies.
Second, for each new task an engineer must hand-write the fitness function, scaffold the harness, and cast the
informal goal into a well-posed, constraint-bound problem grounded in the platform, a hard, expert-intensive step
left outside the automation.

\vspace{2pt} \noindent \textbf{Requirements.}
These shortcomings reduce to three missing ingredients.
Prior agentic systems are not \emph{formal}: they assume a hand-crafted fitness function and per-task scaffolding
instead of a structured, searchable problem grounded in the target's real workload and platform, so they optimize for
correctness or a single objective.
They are not \emph{transferable}: each effort is bespoke, so nothing learned on one task carries to the next.
Finally, they are not \emph{systematic}: the LLM is their only source of candidates, so the search is narrow, biased, and prone to
local optima.
Closing this gap calls for a framework that is all three: \emph{formal}, \emph{transferable}, and \emph{systematic}. 

\section{\system{} Overview}
\label{sec:overview}

\begin{figure*}[t]
  \centering
  \includegraphics[width=\textwidth]{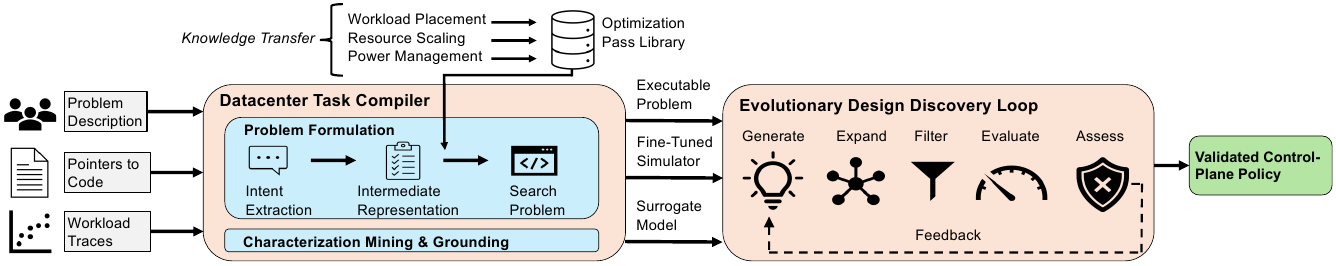}
  \vspace{-6mm}
  \caption{Overview of \system{}.
  The \compiler{} \emph{formalizes} an informal request into an IR and lowers it to an
  executable search problem.
  The \loopAgent{} \emph{searches} that problem and returns a validated policy.
  Structured feedback closes the loop. 
  }
  \label{fig:overview}
  \vspace{-4mm}
\end{figure*}

\system{} turns a plain-language policy request into a validated control-plane policy through a single pipeline with two components (\Cref{fig:overview}). 
The \emph{\compiler{}} is the front-end:
it \emph{formalizes} the request into a machine-checkable \emph{intermediate
representation (IR)}, then lowers this \emph{transferable} IR to an executable search problem (\Cref{sec:compiler}).
The \emph{\loopAgent{}} is the back-end:
it \emph{systematically searches} that problem and returns a policy that satisfies the IR
(\Cref{sec:loop}).
Structured feedback from the loop closes the cycle, steering the compiler's next round.

\vspace{2pt} \noindent \textbf{Running example.}
We use a  control-plane policy request as a running example: \emph{``minimize a service's resource usage without increasing latency.''}
This is a resource-scaling task for a service with a tail-latency SLO (e.g., p99 $\le 50$\,ms), running on a fixed server platform and accompanied by historical load traces.
The request is intentionally underspecified.
The \compiler{} translates it into a \emph{formal}, executable policy specification (\Cref{sec:compiler}), while the \loopAgent{} \emph{systematically} searches the policy space for a solution that satisfies it (\Cref{sec:loop}).

\section{\compiler{}}
\label{sec:compiler} 

The \compiler{} is the front-end of \system{}.
Its purpose is to transform an engineer's intent into an executable search problem.
Existing agentic systems typically embed problem definitions inside prompts, fitness functions, or handwritten evaluation code.
As a result, objectives, constraints, and assumptions remain implicit, knowledge learned solving one task is difficult to reuse on the next, and search has little structure beyond what is encoded in the LLM itself.

The \compiler{} addresses these limitations through three principles. 
First, it makes policy problems \emph{formal} by translating natural-language requests into explicit specifications of decision variables, objectives, constraints, evaluation criteria, and execution requirements. 
Second, it makes optimization \emph{transferable} by representing reusable control knowledge independently of any particular domain. 
Third, it lets the \loopAgent{} search \emph{systematically} by lowering the resulting specification into a well-structured optimization problem. 

\begin{figure}[t]
  \centering
  \includegraphics[width=\columnwidth]{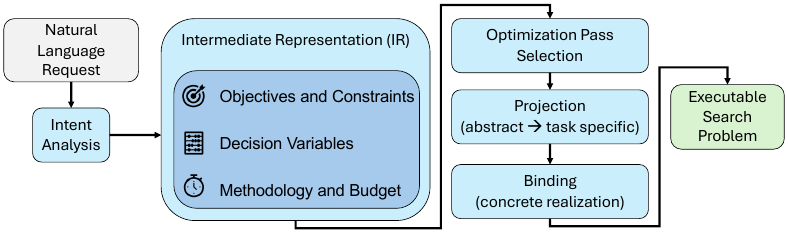}
  \vspace{-6mm}
  \caption{
  The \compiler{} raises a natural-language request into a formal IR, then lowers it into an executable search problem for the \loopAgent{}.} 
  \label{fig:compiler}
  \vspace{-6mm}
\end{figure}

To achieve these goals, the \compiler{} performs three steps (\Cref{fig:compiler}).
It first raises a natural-language request into a uniform intermediate representation (IR) describing what the policy should accomplish.
It then augments the IR with reusable control knowledge selected from a shared library.
Finally, it lowers the resulting specification into an executable search problem comprising a parameter space, candidate generator, evaluation stack, and execution constraints.
The same compilation pipeline supports different classes of control-plane policies while allowing both knowledge and evidence to accumulate across tasks.

\subsection{Raising Intent into a Formal Specification}
\label{sec:compiler-raise}

The compiler's primary input is a natural-language policy request.
Engineers may additionally provide supporting artifacts such as workload traces, source code, deployment specifications, historical configuration changes, or reference designs.
The goal of the raising stage is to transform these inputs into a \emph{formal} representation of the policy problem.
The raising stage performs two tasks. 

\vspace{2pt} \noindent \textbf{Interpreting intent.}
The compiler first \emph{interprets intent}.
Agentic components identify the control decisions exposed by the target system, the optimization objectives implied by the request, and the constraints that must be respected.
For the running example, \emph{``minimize a service's resource usage without increasing latency''}, resource usage becomes the optimization objective while latency becomes a hard constraint.

A \emph{critic} then checks the draft.
The critic is a rule-based validator, not another LLM, so its checks are deterministic and repeatable.
It rejects anything the request does not support (e.g., a constraint left implicit, a number the request never
stated, or a guessed workload size) and re-prompts the agents to repair the draft.
This repeats until the IR is well-formed, so the compiler never passes off an invented value as an
engineer-provided fact.
Gaps the request leaves open are filled from domain playbooks (e.g., the default SLO metric and evaluation
methodology conventional for scaling policies) so a terse request still yields a complete problem.

The critic enforces the IR's typing rules: decision variables declare type and domain, objectives and constraints refer to declared metrics bound to an evaluation source, and constraints are machine-checkable predicates over observable quantities.
The critic rejects undefined variables, inconsistent units, duplicate bindings, and constraints it cannot evaluate: every accepted IR is well-typed and machine-checkable. 

\vspace{2pt} \noindent \textbf{Mining characterization.}
The front-end then \emph{mines workload and platform characterization data} to ground the IR in reality.
Rather than reasoning about the service in the abstract, it retrieves the artifacts that make the problem concrete: historical load traces, latency response under different resource allocations, and relevant platform constraints.
This characterization is attached to the IR and later drives the simulator and surrogate generated during lowering (\Cref{sec:compiler-lower}).
As a result, the search optimizes for the target workload and hardware rather than a generic proxy.

\subsection{Intermediate Representation}
\label{sec:compiler-ir}

The result of the raising-intent stage is a uniform intermediate representation (IR).
The IR serves as the contract between the compiler and search loop:
regardless of domain, every control-plane problem is represented using the same structure.
It is a validated schema and not a free-form text, so the back-end can reason over its fields without re-interpreting a prompt during search. 
\Cref{lst:ir-example} shows the IR for the running example. We describe its parts in turn.

\vspace{2pt} \noindent \emph{Decision variables.}
These name what the policy controls and the space it may choose from.
For the running example, they are the per-interval replica count and the CPU and memory limits, each with a type
and a valid domain.
The decision variables define the search space the back-end explores.

\vspace{2pt} \noindent \emph{Objectives.}
These state what to optimize and in which direction (e.g., minimize allocated resources).
A problem may carry several competing objectives.
The IR records each objective's direction and relative priority, so the back-end can reason about trade-offs and not collapse them into a single scalar.

\vspace{2pt} \noindent \emph{Constraints.}
These are hard predicates a policy must never violate (e.g., the latency SLO, capacity limits, or power envelopes).
Constraints are what separate control-plane problems from single-objective optimization.
A candidate that wins on the objective but breaks a constraint is rejected outright.

\vspace{2pt} \noindent \emph{Evaluation methodology.}
This specifies how a candidate is scored.
It names the simulator that computes outcomes, the workload trace that drives it, and the metrics read back.
Fixing evaluation in the IR makes results comparable across candidates and reproducible across runs.

\vspace{2pt} \noindent \emph{Execution budget.}
This records how much time the policy has to act (e.g., tens of milliseconds for a placement decision or a few seconds for a scaling controller).
The budget is a first-class part of the IR because it shapes the policy design itself:
tighter budgets require cheaper decision procedures, while looser budgets allow more extensive search and evaluation.
The compiler carries this budget through lowering to ensure the generated policy satisfies its runtime constraints.

\begin{lstlisting}[style=ir,float=t,captionpos=b,
  caption={The IR the \compiler{} emits for the running example,
    \emph{``minimize a service's resource usage without increasing latency.''}},
  label={lst:ir-example}]
task_type: resource scaling
decision_variables:
  replicas   : int    in [1, 12]
  cpu_limit  : float  in [0.5, 4.0]  # cores
  mem_limit  : float  in [0.5, 8.0]  # GB
objectives:
  minimize allocated_resources       # priority 1.0
constraints:                         # hard
  p99_latency_ms <= 50
  cooldown >= 1 interval
execution_budget:
  class=loose  interval=30s  mode=full
evaluation:
  simulator = autoscaling_evaluator
  trace     = mined_load(service)
  metrics   = {p99_latency, resources,
               slo_violation_rate}
characterization:                    # grounding
  load profile, latency-vs-alloc curve,
  platform limits
\end{lstlisting}

\subsection{Transferable Control Knowledge}
\label{sec:compiler-transfer}

A formal specification describes \emph{what} should be optimized, but not \emph{how} to solve the problem.
Solving every policy request from scratch would discard the valuable knowledge accumulated from previous tasks.
The compiler therefore maintains a shared library of reusable, domain-agnostic control knowledge.
Then, given an IR, the compiler selects the relevant pieces and projects each onto the target domain.

\vspace{2pt} \noindent \textbf{The pass library.}
The library is organized around \emph{optimization passes}.
An optimization pass is a reusable, domain-agnostic control idea.
For example, a capacity-safety pass stops a policy from ever committing more resources than a machine has, and a
burst-prediction pass reacts to an early demand signal so a service scales out before its latency suffers.
Passes live in the shared library, one entry per pass.
An entry is a structured record with four parts: an \emph{applicability} condition, the \emph{obligations} the pass is meant to meet, a \emph{lowering} for each target domain, and a \emph{confidence} learned from past runs. 
The applicability condition is a predicate over IR features (e.g., the workload is bursty and a latency SLO is
present).
An obligation is a requirement the final policy must satisfy (e.g., respect a hard capacity limit or a latency SLO).

\vspace{2pt} \noindent \textbf{Selecting passes.}
The compiler turns the IR into a set of passes in three steps.
First, \emph{eligibility}: the compiler keeps a pass only when the IR matches the pass's applicability condition and the pass has a lowering for the target domain, and it drops any pass that past evidence has flagged as harmful here.
Second, \emph{scoring}: the compiler ranks each filtered pass by how well the IR's features (e.g., its objectives and constraints) overlap the pass's descriptors.
Third, \emph{obligation cover}: the compiler treats hard constraints and objectives in the IR as 
obligations, then greedily picks the smallest, highest-scoring set of passes that together meet all of them, breaking ties by confidence.
The result is a small, well-supported set of passes chosen so that every constraint and objective is 
addressed by at least one of them.
This library is \system{}'s encoded expertise: a growing, evidence-backed store of control ideas that the compiler reuses here and that also seeds the loop's generation (\Cref{sec:loop}).

\begin{figure}[t]
\vspace{-4mm}
  \centering
  \includegraphics[width=\linewidth]{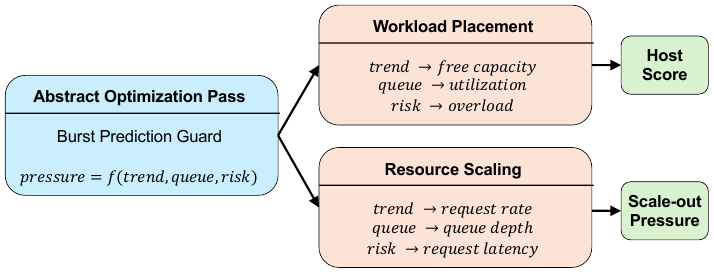}
  \vspace{-7mm}
  \caption{Projection as knowledge transfer.
  One abstract pass carries a canonical formula that projection
  re-instantiates per domain. 
  }
  \label{fig:projection}
  \vspace{-6mm}
\end{figure}

\vspace{2pt} \noindent \textbf{Projecting passes.}
A selected pass names an idea but not its concrete form in the target.
Thus, each entry has a \emph{canonical formula}: a target-neutral expression over abstract variables (e.g., demand trend, queue pressure, and SLO risk).
Projection specializes this formula to a target in two steps.
It first binds each abstract variable to a concrete field the target exposes.
It then emits the bound expression as a target-native term (e.g., a host-score term for placement or a controller
term for scaling).
\Cref{fig:projection} shows 
an example of this binding. 

Consider the burst-prediction guard, whose canonical formula is a short-horizon demand pressure,
$\mathit{pressure} = f(\mathit{trend}, \mathit{queue}, \mathit{slo\_risk})$.
For resource scaling, projection binds $\mathit{trend}$, $\mathit{queue}$, and $\mathit{slo\_risk}$ to the service
signals $\mathit{request\_rate\_trend}$, $\mathit{queue\_depth}$, and $\mathit{latency\_slo\_ratio}$, and emits
their combination as a scale-out pressure term.
For workload placement, projection binds the same three variables to host signals (i.e., free headroom,
post-placement slack, and oversubscription risk), and emits a headroom-reserving host score.
The idea transfers, but the concrete term does not, so a guard proven for scaling need not be re-derived for
placement.
When a projection needs a field the target does not expose, the compiler proposes a
new primitive or falls back (\Cref{sec:compiler-formula}).

\subsection{Lowering to an Executable Search Problem}
\label{sec:compiler-lower}

After pass selection and projection, the compiler has a formal problem definition and a set of relevant reusable control ideas. 
The final stage lowers this representation into an executable search problem consumed by \loopAgent{}.
Lowering produces four artifacts:
a parameter space, a generator, a surrogate, and a simulator.

The \emph{parameter space} is the concrete search space: each decision variable and each pass parameter becomes a
typed knob with a range and a default (e.g., a per-interval replica count in $[1,12]$ or a burst-pressure threshold).
The \emph{generator} proposes candidate policies over this space, seeded with the selected passes so its first
candidates reflect established practice. 
The \emph{surrogate} and the \emph{simulator} score those candidates. 

\vspace{2pt} \noindent \textbf{The evaluation stack.}
The surrogate and simulator form the problem's \emph{evaluation stack}. 
The compiler \emph{defines} this stack by producing a surrogate model specification and augmenting the high-fidelity simulator named in the IR.

For the surrogate, the compiler determines the inputs, outputs, and target metrics from the IR and the mined characterization data.
The Evolutionary Design Discovery Loop then instantiates and continually refits this surrogate using measured evaluation results (\Cref{sec:loop-filter}), allowing it to rank many candidates before paying for full simulation.

For the simulator, the compiler configures the evaluation environment using the mined characterization data so scores reflect the target workload and platform.
When the IR requires a signal the simulator does not yet expose (e.g., a new tail-latency percentile or a power-headroom metric), the compiler extends the simulator to emit that signal.

The surrogate keeps broad search affordable, while the augmented simulator keeps evaluation faithful to the target system.
The compiler shapes the search problem to the IR's execution budget, ensuring the generated policy satisfies its runtime constraints.
\Cref{alg:compile} summarizes the procedure.

\begin{algorithm}[t]
\caption{Compiling a request to a search problem}
\label{alg:compile}
\begin{algorithmic}[1]
\REQUIRE nat.-lang. request $r$, target domain $d$, pass libr. $L$
\STATE $\mathit{ir} \gets \textsc{Raise}(r)$ \COMMENT{interpret intent, mine characterization}
\STATE normalize and validate $\mathit{ir}$; fill gaps from playbooks
\STATE $P \gets \textsc{SelectPasses}(\mathit{ir}, L)$ \COMMENT{eligibility, semantic score, obligation cover}
\FORALL{pass $p \in P$}
  \STATE $t_p \gets \textsc{Project}(p, d)$ \COMMENT{can. form. $\rightarrow$ target-native term}
\ENDFOR
\FORALL{control idea required by $\mathit{ir}$ but absent from $P$}
  \STATE propose a safe Formula-DSL primitive
  \STATE register it if the target exposes its fields, else fall back
\ENDFOR
\STATE $\mathit{prob} \gets \textsc{Bind}(\mathit{ir}, \{t_p\})$ \COMMENT{generator, surrogate, simulator,
parameter space}
\STATE shape $\mathit{prob}$ to $\mathit{ir}$'s execution budget
\RETURN executable search problem $\mathit{prob}$
\end{algorithmic}
\end{algorithm}

\vspace{-5mm}
\subsection{Open-World Extensions}
\label{sec:compiler-formula}

The \emph{pass library} cannot anticipate every policy. 
New problems may require control concepts not captured by existing passes. 
The compiler proposes the missing concept as a new primitive. 
Primitives are expressed in a safe \emph{formula DSL}: a restricted expression tree over target fields using only approved arithmetic operators and no code execution.
Because primitives describe declarative control logic rather than executable code, the compiler can analyze and validate them against the target's capabilities before incorporating them.

If the required fields are available, the primitive is registered and reused in future compilations.
Otherwise, the compiler reports the capability gap and falls back to the closest supported pass. 
This design preserves safety while enabling extensibility.

\subsection{Evidence-Guided Library Evolution}
\label{sec:compiler-evolution}

The shared IR and library let optimization knowledge compound across tasks.
Projection turns an idea learned in one domain into a native IR construct, allowing a pass refined for one problem to improve others.
Onboarding a new domain therefore \emph{extends} the library rather than starting from scratch:
most domains reuse existing passes, while new domains contribute passes that future problems inherit.

The library also learns from evidence.
Each pass maintains a confidence score that is updated from measured outcomes, and evidence of negative transfer progressively suppresses passes that perform poorly for a given domain.
Over time, the library evolves from a collection of expert-written control ideas into an evidence-backed repository of where and when they work, with consistently effective passes selected more aggressively.

This accumulation of \emph{formal} specifications, \emph{transferable} control knowledge, and empirical evidence is the mechanism behind \system{}'s generality. 

\section{\loopAgent{}}
\label{sec:loop}

The \loopAgent{} is \system{}'s back-end.
It takes the executable search problem produced by the \compiler{} and searches for a policy that satisfies the IR.
It supplies the third ingredient: it makes policy generation \emph{systematic}.

A single LLM proposal explores only a narrow region of the design space and often converges to a local optimum (\Cref{sec:background}).
The loop overcomes this limitation in two ways.
Within each round, it expands the search far beyond a single proposal, deriving many structural and parametric variants from each seed policy.
Across rounds, it improves iteratively, incorporating feedback from previous evaluations into subsequent exploration. A policy is accepted only after it is validated against the IR's constraints.

\begin{figure}[t]
  \centering
  \includegraphics[width=\columnwidth]{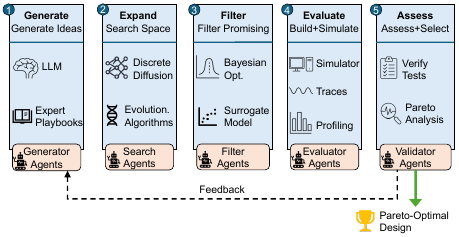}
  \vspace{-6mm}
  \caption{
  Each round of the \loopAgent{} expands a few seeds into many variants, filters them with a surrogate, evaluates the retained few in high fidelity, and feeds the results back to steer the next round. 
  }
  \label{fig:loop}
  \vspace{-5mm}
\end{figure}

Each round consists of five stages:
\emph{generate}, \emph{expand}, \emph{filter}, \emph{evaluate}, and \emph{feedback}.
\Cref{fig:loop} shows these stages form a closed loop: a small set of seed policies expands into many candidates, a low-cost surrogate filters the search space, the most promising candidates are evaluated in high fidelity, and the resulting feedback guides the next round.

\subsection{Generate}
\label{sec:loop-generate}

Each round begins by \emph{generating} seed policies.
LLM agents propose candidates over the IR's decision variables, grounded in the workload and platform
characterization the \compiler{} attached.
The agents are not left to guess: they draw on the same shared library the compiler uses (i.e., reusable primitives, whole-policy templates, past failure episodes, and a map from workload signals to policies that worked before).
Seeded with this distilled expertise, the first candidates reflect established practice for the policy type (e.g., for the running example, familiar scaling strategies such as load-proportional or headroom-based allocation) expressed as executable policy logic.
An agent may also propose a control idea the library lacks, which the compiler admits as a new primitive
(\Cref{sec:compiler-formula}).
Seeds are generated independently, so a malformed proposal never stalls the round.

\subsection{Expand}
\label{sec:loop-expand} 

\begin{figure}[t]
  \centering
  \includegraphics[width=\columnwidth]{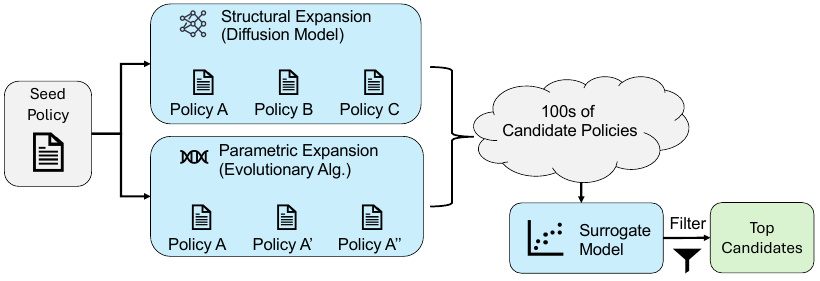}
  \vspace{-6mm}
  \caption{Expansion and filtering. A seed is expanded along two axes, i.e., \emph{structural} variants from the
  diffusion step and \emph{parametric} variants from the evolutionary search, producing many candidates. The
  surrogate ranks them cheaply, and only a small, high-value subset is simulated in high fidelity.}
  \label{fig:loop-search}
  \vspace{-5mm}
\end{figure}

A handful of seeds cannot cover the design space, so the loop \emph{expands}  along two axes.
\Cref{fig:loop-search} shows the axes:
one changes a policy's \emph{structure}, the other tunes its \emph{parameters}.

\vspace{2pt}
\noindent \textbf{Structural search.}
The loop explores structure with a \emph{discrete-diffusion} step.
It masks one part of a policy (e.g., scoring logic, execution mechanism, safety guards) and
regenerates only that part while copying the rest verbatim.
Masking one part at a time changes a policy's structure without disturbing the elements already known to work (e.g., it can rewrite how a scaling policy reacts to a load spike while leaving its SLO guard intact).
The step is fail-closed: a regeneration that touches anything outside its mask is rejected and retried. 

At the implementation level, the loop realizes each diffusion step with an LLM: the LLM performs the mask-and-denoise step, and we do not train a dedicated diffusion model.
It hands the model the parent policy, the mask that marks the one editable region, and the instruction to copy
everything outside the mask unchanged.
The masks form a small fixed set, one per structural facet, so a step is always a bounded, interpretable edit.
The system compares the child against the parent and rejects it if anything outside the mask moved,
re-prompting until the edit is clean or falling back to a deterministic structural edit.
To recombine ideas that already work, the loop can condition the step on the current best policies, so the
regenerated region fuses their strong features. 

\vspace{2pt} \noindent \textbf{Parametric search.}
For each structure, an evolutionary search tunes the free parameters (e.g.,
thresholds, window sizes, and headroom margins).
Mutation perturbs weights and numeric knobs around a parent, and a surrogate-guided simulated-annealing step takes local moves, occasionally accepting a worse candidate to escape a local optimum.

At the implementation level, this is a small evolutionary loop over a population of policies.
Mutation samples multiplicative Gaussian noise on each term's weight and on the numeric knobs, and nudges integer
knobs by a step, producing children clustered around a promising parent.
The annealing step runs a short chain per parent and scores each proposed move with the surrogate rather than the
simulator, so it can take many cheap local steps before anything is simulated.
Successive generations draw their parents from the measured Pareto frontier, so parametric search compounds across
rounds instead of restarting.
Structural search thus explores qualitatively different policies, while parametric search sharpens each one.
Together they push the search past the single region an LLM tends to fixate on, and they quickly produce far more
candidates than can be simulated.

\subsection{Filter}
\label{sec:loop-filter}

Expansion produces far more candidates than the loop can afford to simulate, so it \emph{filters} the pool with a
cheap \emph{surrogate} before paying for high-fidelity evaluation.
The surrogate is a fast model of the simulator: for a given candidate on a given workload regime, it predicts the
bundle of metrics the IR scores, e.g., SLO attainment, resource cost, and overload risk, each with a confidence
rather than a single point estimate.

\vspace{2pt} \noindent \textbf{Exploration-aware ranking.}
The loop turns these predictions into one \emph{ranking score} that is deliberately exploration-aware.
It rewards high predicted utility, but it also adds a bonus for uncertain candidates so that informative designs are not discarded: a bonus for candidates that would repair a known failure, and a bonus for novelty, while penalizing candidates that look strong only where it has little data.
Ranking by this score, the loop keeps a small, high-value subset for full evaluation and drops the rest.
This is what makes broad expansion affordable: the loop can propose orders of magnitude more candidates than it
could ever simulate, and it spends simulation only on the few most worth it.

\vspace{2pt} \noindent \textbf{Surrogate model.}
The surrogate is produced by the \compiler{} during lowering; the loop fits and uses it. 
At the implementation level, the surrogate is an uncertainty-aware regressor trained on the growing table of past
simulator results, with one model per target metric, e.g., an ensemble of a Gaussian process and tree-based
regressors whose disagreement supplies the confidence.
To score a policy it has never run, it encodes the candidate into a feature vector: which score terms the policy
uses and with what weights, its execution mechanism and knobs, fingerprints of its formulas taken from the
compiler's DSL, and a behavioral signature obtained by running the policy on a few tiny synthetic probes and
recording its choices; the workload regime is encoded alongside.
This encoding is what lets even a brand-new policy receive a grounded prior instead of a blind guess.
A local calibration step then pulls each prediction toward its nearest measured neighbors and dampens
over-optimistic scores far from data, so the filter does not chase mirages. 

\vspace{2pt} \noindent \textbf{Quota-based selection.}
Selection is quota-based: the loop fills most slots with the best-ranked candidates but
reserves slots for failure repairs, alternative mechanisms, high-uncertainty audits, and novel designs.
Thus, a promising idea is not starved by many near-duplicates of the current best.

\vspace{2pt} \noindent \textbf{Trust and refitting.}
The surrogate is trusted when enough measured data has accumulated (e.g., a minimum number of distinct policies and workloads). Until then the loop screens conservatively.
Each round refits the surrogate on the newly measured results, so the filter sharpens over time.

\begin{figure*}[t]
  \centering
  \includegraphics[width=\textwidth]{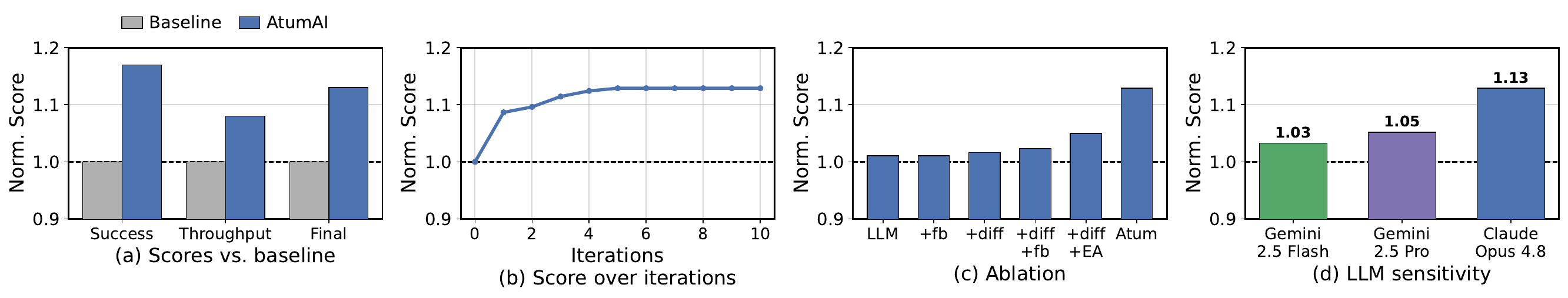}
  \vspace{-8mm}
  \caption{Workload placement. (a) Baseline vs.\ \system{} per quality component and final score, normalized to the
  baseline ($=1.0$, dashed);
  (b) best score per generation;
  (c) ablation of the loop's mechanisms;
  (d) sensitivity
  to the loop's LLM. 
  }
  \label{fig:eval-vm}
  \vspace{-4mm}
\end{figure*}

\subsection{Evaluate and Validate}
\label{sec:loop-evaluate}

The retained candidates undergo deep \emph{evaluation} in the high-fidelity simulator generated during lowering (\Cref{sec:compiler-lower}) and driven by the mined workload and platform characterization.
As a result, measurements reflect the target workload and hardware rather than an abstract model.

Importantly, the loop evaluates each candidate across a \emph{matrix} of workload regimes (e.g., steady, bursty, and fragmented load), not a single trace, exposing policies that overfit specific conditions.
Evaluation computes objective values \emph{and} checks every IR constraint.
A candidate is \emph{accepted} only if it satisfies all constraints; a policy that improves the objective but violates a constraint (e.g., reduces resource usage while breaching a tail-latency SLO) is rejected.

Two safeguards enforce this property.
First, a feasibility audit fails closed on candidates that violate resource or execution budgets before simulation.
Second, an offline oracle distinguishes avoidable inefficiency from the workload's inherent limits.
The loop also continuously validates the surrogate against simulator results (e.g., using rank correlation and selection regret) to ensure the low-cost filter remains accurate as the search progresses.

This stage grounds optimization in high-fidelity measurements.
It ensures \system{} returns policies that satisfy constraints across diverse operating conditions.

\subsection{Feedback and Iteration}
\label{sec:loop-feedback}

Finally, the loop turns results into \emph{structured feedback}.
It builds a failure summary (i.e., which constraint failed, on which regime, and by how much) and a per-candidate brief that pairs each promising policy with a stronger comparator to learn from. 
This feedback steers the next round: it re-seeds generation toward the observed failures, aims the diffusion mask at the weak part of a policy, and picks the next round's parents from the measured Pareto frontier. 
Winners and the worst candidate are written back to the shared library as positive and negative examples, so expertise accumulates across rounds and carries to future problems (\Cref{sec:compiler-transfer}).
\Cref{alg:loop} summarizes one round.
The loop runs for a fixed budget of rounds and returns the best validated policy it found.

\begin{algorithm}[t]
\caption{One round of the Evol. Design Discovery Loop}
\label{alg:loop}
\begin{algorithmic}[1]
\REQUIRE search problem $S$, parents $P$, surrogate $M$, feedback $B$, library $\mathcal{L}$
\STATE $\mathit{seeds} \gets \textsc{Generate}(S, \mathcal{L}, B)$ \COMMENT{playbook-guided LLM seeds}
\STATE $\mathit{cand} \gets \textsc{Diffuse}(P \cup \mathit{seeds}) \cup \textsc{Mutate}(P \cup \mathit{seeds})
\cup \textsc{Anneal}(P \cup \mathit{seeds}, M)$
\STATE $\mathit{pool} \gets P \cup \mathit{seeds} \cup \mathit{cand}$
\STATE $\mathit{ranked} \gets \textsc{Rank}(\mathit{pool}, M)$ \COMMENT{uncertainty-aware surrogate}
\STATE $\mathit{filter} \gets \textsc{Select}(\mathit{ranked})$ \COMMENT{small quota for simulation}
\STATE $\mathit{rows} \gets \textsc{Simulate}(\mathit{filter})$ over workload regimes; drop constraint violators
\STATE $M \gets \textsc{Refit}(M, \mathit{rows})$
\STATE $P \gets \textsc{ParetoParents}(\mathit{rows})$;\; $B \gets \textsc{Failures}(\mathit{rows})$
\STATE $\mathcal{L} \gets \mathcal{L} \cup \{\text{best}, \text{worst}\}$ \COMMENT{positive and negative examples}
\RETURN best validated policy in $\mathit{rows}$
\end{algorithmic}
\end{algorithm}

\section{Use-Case Studies}
\label{sec:use-cases}

We evaluate \system{} on the three control-plane use cases from \Cref{sec:overview}.
Each subsection first states the problem, then the methodology, and then reports results.
Unless noted, the \loopAgent{} uses Claude Opus 4.8 as its LLM and every reported policy is measured in the
high-fidelity simulator, not the surrogate.
Each simulator is validated against a real system (e.g., we cross-check the resource-scaling simulator against a
Kubernetes deployment~\cite{k8s}) so the reported gains reflect deployable behavior. 
All scores are normalized to their use case's expert-engineered baseline (i.e., the Gen-0 policy an operator ships today) so they are meaningful within a use case but not comparable across use cases.
For each use case we report four studies:
the final policy against the baseline,
the score over generations,
an ablation of the loop's mechanisms,
and sensitivity to the LLM.
The ablation's LLM-plus-feedback configuration emulates state-of-the-art LLM-driven evolutionary systems~\cite{novikov2025alphaevolvecodingagentscientific,skydiscover}, and serves as an upper bound on them, as these systems lack the formal problem formulation the \compiler{} provides. 

\subsection{Workload Placement}
\label{sec:eval-vmplacement}

\vspace{2pt} \noindent \textbf{Problem.}
The objective is to maximize admission rate and scheduler throughput while ensuring colocated VMs run without contention.
Each decision must satisfy two constraints:
the selected host must have sufficient free CPU and memory and colocation policies may prohibit certain VMs from sharing a host.
The decision involves a trade-off: packing VMs more aggressively can increase admissions but may create CPU hotspots that degrade the performance of existing VMs.

\vspace{2pt} \noindent \textbf{Methodology.}
We simulate a cluster with 90 hosts, each with 32 CPU cores and 384\,GiB of memory.
The load is a stream of VM arrivals driven by Azure VM traces~\cite{azureTracesV2}.
The baseline is a hand-tuned best-fit packing policy of the kind production cluster managers use~\cite{borg,protean}.
\system{} searches for a placement policy over tens of generations, scoring each candidate with \emph{VMScore}, a
weighted sum of admission success (0.50), planning throughput (0.20), and CPU tail-hotspot penalties (0.20 for the P50, 0.10 for the P99), normalized to the baseline.

\vspace{2pt} \noindent \textbf{Results.}
\Cref{fig:eval-vm}(a) compares the baseline and the \system{} policy across VMScore's three quality metrics and overall score, normalized to the baseline.
\system{} achieves a $1.13\times$ higher overall score, improving admission success by $17\%$ and planning throughput by $8\%$ without increasing CPU hotspots.

The gain comes from complementary improvements in placement quality and scheduler efficiency.
The policy's admission logic combines harmonic CPU/memory headroom, minimum-slack placement, and oversubscription guards to preserve more usable residual capacity than the baseline's best-fit heuristic, allowing more future arrivals to be admitted.
To improve throughput, it avoids exhaustive host scoring by grouping machines by topology, ranking groups with a lightweight headroom proxy, and evaluating only a bounded subset of hosts from the most promising group.
Scoring fewer machines 
reduces scheduling overhead and increases throughput.
Together, these mechanisms improve both admission success and scheduler throughput while maintaining the same level of contention and tail overload.

\vspace{2pt} \noindent \emph{Discovery path.} The capacity-safety and oversubscription guards were transferred from the shared library, whereas the harmonic CPU/memory headroom scoring, minimum-slack placement, and topology-grouped search were discovered by the loop and then tuned by feedback.

\begin{figure*}[t]
  \centering
  \includegraphics[width=\textwidth]{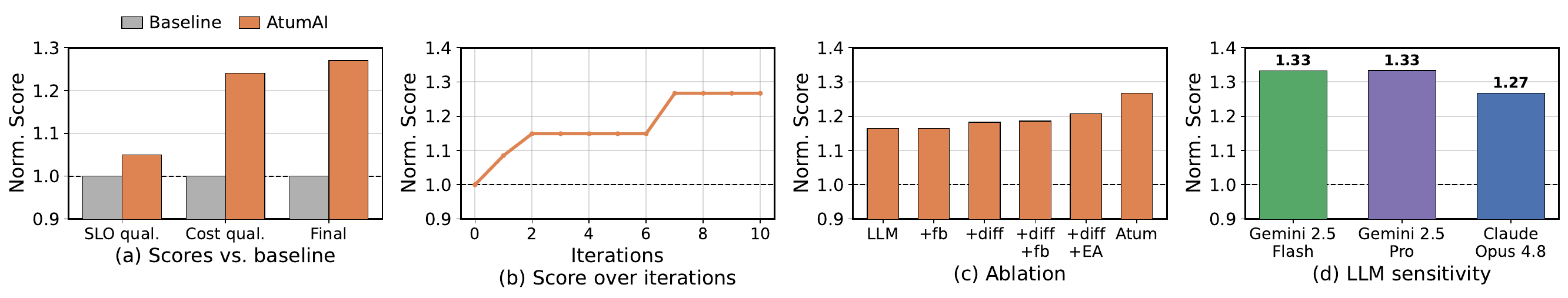}
  \vspace{-8mm}
  \caption{Resource scaling.
  (a) Baseline vs.\ \system{} per quality component and final score, normalized to the
  baseline ($=1.0$, dashed);
  (b) best score per generation;
  (c) ablation of the loop's mechanisms;
  (d) sensitivity
  to the loop's LLM.}
  \label{fig:eval-scaling}
  \vspace{-5mm}
\end{figure*}

\vspace{2pt} \noindent \emph{Loop iteration.}
\Cref{fig:eval-vm}(b) shows the best validated score over generations.
The first generation outperforms the baseline, as the playbook-seeded admission policy is strong out of the box.
Subsequent gains come from evolutionary mutations and annealing steps that reduce per-placement scoring cost and improve throughput.
Each generation produces roughly 130 candidates, of which the surrogate forwards seven for detailed simulation.
The score reaches $1.13\times$ by generation~5 and then saturates, with no subsequent variant further improving.

\vspace{2pt} \noindent \emph{Mechanism ablation.}
\Cref{fig:eval-vm}(c) shows an ablation that enables the loop's mechanisms one at a time.
The mechanisms are worth little in isolation: the LLM alone reaches $1.01\times$, adding diffusion
and evolutionary search raises it only to $1.05\times$, and feedback on its own adds negligible gains.
Feedback is instead a multiplier: layered on top of diffusion and search, it lifts the score to $1.13\times$.
The reason is that diffusion and search generate structural and parametric variants but explore blind, whereas
feedback identifies the measured bottleneck (i.e., the per-placement scoring cost) and steers the next variants to attack it.
Closing the loop converts a pool of variants into the mechanism change that wins.

\vspace{2pt} \noindent \emph{LLM sensitivity.}
\Cref{fig:eval-vm}(d) shows that every LLM discovers a policy that outperforms the baseline, indicating that most of the gain comes from \system{}'s compiler and search loop rather than the LLM.
Claude Opus 4.8 achieves the highest score at $1.13\times$, while Gemini~2.5 Pro and Flash reach $1.05\times$ and $1.03\times$, respectively.
More capable LLMs raise the ceiling, but \system{} consistently finds strong policies across models.

\subsection{Resource Scaling}
\label{sec:eval-scaling}

\vspace{2pt} \noindent \textbf{Problem.}
Resource scaling decides how much CPU and memory each service receives over time.
The objective is to minimize resource allocation while keeping every service within its latency SLO.
The hard constraint is finite cluster capacity: the combined allocations of all services must fit within the available CPU and memory.
The decision is inherently a trade-off: overprovisioning preserves SLOs but wastes capacity, while aggressive consolidation improves efficiency but risks violations during load spikes.

\vspace{2pt} \noindent \textbf{Methodology.}
We simulate 100 services that share 30 nodes.
The load spans 11 workload regimes driven by production traces from Alibaba~\cite{alibabaTraces} (e.g., steady, bursty, trending, and dependency-driven traffic), so a policy cannot overfit one pattern.
The baseline is a reactive, threshold-based autoscaler of the kind used in production~\cite{autopilot}.
\system{} searches for a scaling controller over tens of generations, scoring each candidate as \emph{$0.5\,$SLO quality $+ 0.5\,$cost quality} normalized to the baseline, where SLO quality falls as the violation rate rises and cost quality falls as allocated resources rise.

\vspace{2pt} \noindent \textbf{Results.}
\Cref{fig:eval-scaling}(a) compares the baseline and the \system{} policy across the two quality components and the final score, normalized to the baseline.
\system{} reaches $1.27\times$.
The gain comes almost entirely from cost efficiency: it improves cost quality by $24\%$ (i.e., uses fewer resources) while keeping the SLO-violation rate at $1.3\%$.

The discovered policy runs lean without violating the SLO by matching each action's prediction horizon to its latency impact.
It extrapolates request-rate growth and uses a long, eight-step horizon for slow horizontal scale-out, but a short, one-to-two-step horizon for vertical resizing, scale-in, and budget reconciliation.
It ranks services by weighted SLO risk and allocates CPU to the riskiest first, subject to the cluster budget and a reserve margin.
It scales services down only when both load and latency pressure are low, avoiding the thrashing common in reactive controllers.
By acting on predicted rather than observed pressure, the policy reclaims resources while absorbing bursts before they become SLO violations.

\vspace{2pt} \noindent \emph{Discovery path.} The seed controller inherited burst prediction and the SLO-safety guard from the library, while the horizon-aware planning that separates scale-out from scale-in, and the risk-ranking weights, were discovered by search and then sharpened by feedback.

\vspace{2pt} \noindent \emph{Loop iteration.}
\Cref{fig:eval-scaling}(b) shows the best validated score as the loop iterates.
The score climbs in discrete steps. 
Each step is the loop repairing the dominant failure mode that feedback surfaced (e.g., first eliminating SLO violations under bursts) then trimming steady-state cost.
The score reaches $1.27\times$ within the first several generations and then saturates.

\vspace{2pt} \noindent \emph{Mechanism ablation.}
\Cref{fig:eval-scaling}(c) shows the same pattern as for workload placement, only sharper: the LLM alone reaches $1.16\times$, adding diffusion and
evolutionary search brings it to $1.21\times$, and feedback on its own adds nothing.
Feedback is again the multiplier, lifting the full loop to $1.27\times$.
The controller becomes horizon-aware only once feedback reports that reactive, observe-then-react scaling is the
failure mode, so the loop must both generate program variants and be told which one removes the measured SLO
breaches.

\begin{figure*}[t]
  \centering
  \includegraphics[width=\textwidth]{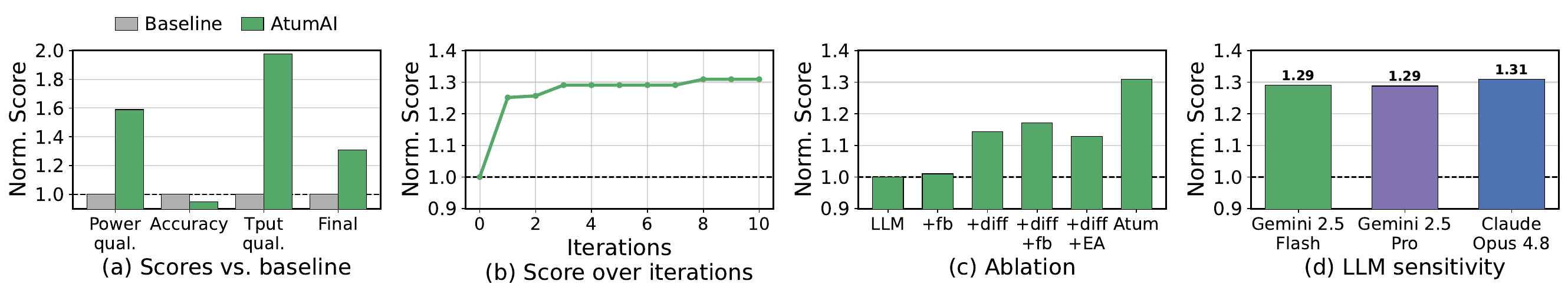}
  \vspace{-8mm}
  \caption{Power management. (a) Baseline vs.\ \system{} per quality component and final score, normalized to the
  baseline ($=1.0$, dashed); (b) best score per generation; (c) ablation of the loop's mechanisms; (d) sensitivity
  to the loop's LLM. 
  }
  \label{fig:eval-power}
  \vspace{-5mm}
\end{figure*}

\vspace{2pt} \noindent \emph{LLM sensitivity.}
\Cref{fig:eval-scaling}(d) shows that
\system{} is robust to this choice: every model exceeds $1.20\times$, the three frontier models
cluster between $1.27$ and $1.33\times$, and even the smallest, Gemini~2.5 Flash-Lite, reaches $1.21\times$.
Here the frontier Gemini models edge out Claude Opus 4.8, the opposite of placement, which reinforces that no single
model is essential: \system{}'s structure sets a high floor, and the best LLM for a task raises the ceiling.

\subsection{Power Management}
\label{sec:eval-power}

\vspace{2pt} \noindent \textbf{Problem.}
Power management allocates a fixed power budget across an LLM-inference fleet.
The objective is to serve as much throughput as possible within the budget while keeping every service accurate.
The hard constraint is a per-service accuracy floor, i.e., each service must retain at least $0.9$ of the accuracy it
would have under the largest model.
The decision is a trade-off, e.g., serving a service with a smaller model saves power and frees throughput but lowers
its accuracy, so it is safe only where the service has accuracy headroom.

\vspace{2pt} \noindent \textbf{Methodology.}
We simulate a 960-server inference fleet over a two-week production trace from Azure~\cite{dynamollm}, where each service can be served by a 7B,
13B, or 70B Llama3 model.
The policy sets, per service, the model size, the request routing, and the server frequency.
The baseline is 
a production fleet power manager~\cite{dynamoFB,thunderbolt}, i.e.,
the highest-accuracy, highest-power option an operator defaults to.
\system{} scores each candidate as
\emph{$0.45\,Q_\mathit{power} + 0.35\,Q_\mathit{accuracy} + 0.20\,Q_\mathit{throughput}$} normalized to the baseline,
and rejects any policy that breaks the accuracy floor for even one service.

\vspace{2pt} \noindent \textbf{Results.}
\Cref{fig:eval-power}(a) shows the baseline and the \system{} policy across the three quality components and the final score, each normalized to the baseline.
\system{} reaches $1.31\times$: it cuts provisioned power by $21\%$ and raises served throughput by
$17\%$, while every service stays above the accuracy floor, the lowest at $0.90$.
The gain is not a blanket switch to smaller models, which would breach the floor.
Instead, the discovered policy assigns a service-specific model mix, dropping a service to a 13B or 7B model only
where it has enough accuracy headroom and keeping 70B elsewhere.
It then routes each request by predicted marginal power, sending load to the server and row that will draw the least
additional dynamic power, and penalizes route churn.
Freeing power on the accuracy-tolerant services is what lets the fleet serve more throughput under the same budget
without violating any constraint.

\vspace{2pt} \noindent \emph{Discovery path.} Only the accuracy-floor enforcement was transferred from the library; the service-specific model mix and the power-aware routing were discovered by the search.

\vspace{2pt} \noindent \emph{Loop iteration.}
\Cref{fig:eval-power}(b) shows that the first generation already captures most of the gain, $1.25\times$, from the service-specific model mix.
The later generations add the power-aware routing refinement, which the loop found through evolutionary mutation and annealing,
reaching $1.31\times$ by generation~8.

\vspace{2pt} \noindent \emph{Mechanism ablation.}
\Cref{fig:eval-power}(c) shows that the LLM alone matches the baseline ($1.00\times$). 
Structural search is the first key ingredient, raising the score to $1.14\times$ by discovering service-specific model mixes.
Feedback provides the largest gain: combined with diffusion and evolutionary search, it lifts the score to $1.31\times$, well above any partial configuration, which reaches at most $1.17\times$.
Diffusion and search generate diverse model-mix and routing policies, while feedback identifies those that best improve throughput per watt without violating the accuracy floor.

\vspace{2pt} \noindent \emph{LLM sensitivity.}
\Cref{fig:eval-power}(d) shows that \system{} is robust to LLM: every model reaches at least $1.28\times$, the three frontier
models shown cluster between $1.29$ and $1.31\times$, and even the smallest, Gemini~2.5 Flash-Lite, reaches
$1.28\times$.
As before, most of the value comes from \system{}'s structure rather than any single LLM, i.e., Claude Opus 4.8 tops the group at $1.31\times$ and a more capable model raises the ceiling only slightly.

\subsection{The Value of the \compiler{}}
\label{sec:eval-compiler}

\begin{figure}[t]
  \centering
  \includegraphics[width=\columnwidth]{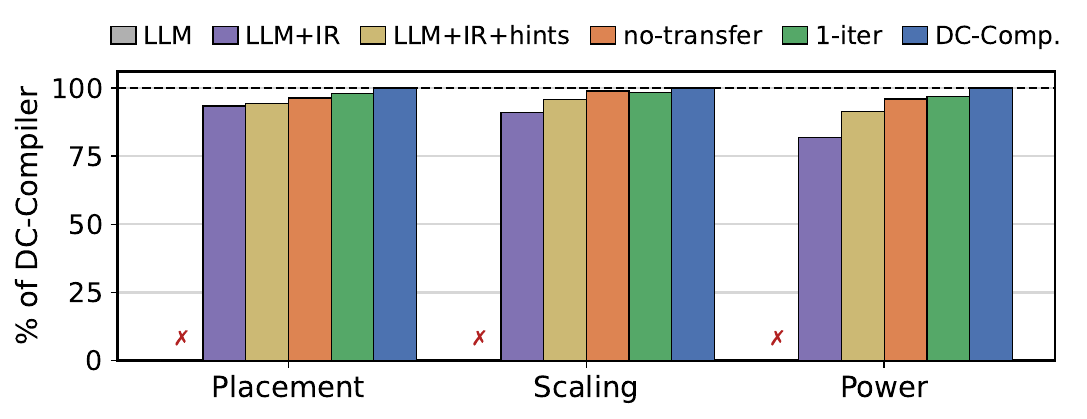}
  \vspace{-7mm}
  \caption{Final policy quality with  six configurations of increasing compiler support
  (an LLM alone, hand-written IR, IR + hints, the \compiler{} without knowledge transfer, a single-iteration
  \compiler{}, and the full \compiler{}), on all three use cases. All scores are normalized to the full \compiler{}; the LLM bars are zero because that configuration produces no valid policy.
  }
  \label{fig:eval-compiler}
  \vspace{-4mm}
\end{figure}

The studies so far isolate the \loopAgent{}; here we isolate the \compiler{}.
\Cref{fig:eval-compiler} shows the final policy quality under six configurations of increasing compiler support,
grouped by use case, with every score expressed as a percentage of the full \compiler{} ($=100\%$).
Partial support is not enough.
An LLM alone fails to emit a valid policy at all, i.e., it never satisfies the hard
constraints, so it scores zero.
Adding a hand-written IR reaches only $82$--$93\%$ of the full system, and adding manual hints on top lifts it to
$92$--$96\%$, but neither closes the gap.
Only the full \compiler{}, which mines characterization and selects passes automatically, reaches $100\%$.
The compiler's formal, grounded IR, not search alone, is what turns a plausible policy into a deployable one.

\vspace{2pt} \noindent \textbf{Knowledge transfer.}
Two further bars isolate individual \compiler{} mechanisms.
Disabling knowledge transfer (i.e., lowering each problem from scratch instead of reusing passes distilled on the other use cases) drops the score to $96.4\%$, $98.9\%$, and $96.0\%$ on placement, scaling, and power.
The drop is largest on power, where reused passes, e.g., capacity-safety and burst-prediction guards, give the
search a strong, feasible starting point instead of a cold start.
A single-iteration \compiler{} that runs each lowering stage once, rather than refining it against feedback, reaches
$98.0\%$, $98.3\%$, and $96.8\%$ on placement, scaling, and power.
Refining the problem formulation makes \system{} robust. 

\section{Conclusion}
\label{sec:conclusion}

Datacenter control-plane policies are increasingly hard to design by hand, and off-the-shelf agentic AI does not fully bridge the gap.
We presented \system{}, a principled framework that generates such policies from a plain-language request: the \compiler{} formalizes the request into a machine-checkable IR, and the \loopAgent{} searches it for a policy that satisfies its objectives and constraints, making policy generation formal, transferable, and systematic.
Across workload placement, resource scaling, and power management, \system{} produced policies that match or exceed expert-engineered baselines from a single framework.

\bibliographystyle{IEEEtranS}
\bibliography{refs}

@inproceedings{micromanycore,
author = {Stojkovic, Jovan and Liu, Chunao and Shahbaz, Muhammad and Torrellas, Josep},
title = {{{\textmu}Manycore: A Cloud-Native CPU for Tail at Scale}},
year = {2023},
booktitle = {Proceedings of the 50th Annual International Symposium on Computer Architecture (ISCA'23)}
}

@inproceedings{deathstarbench,
author = {Gan, Yu and Zhang, Yanqi and Cheng, Dailun and Shetty, Ankitha and Rathi, Priyal and Katarki, Nayan and Bruno, Ariana and Hu, Justin and Ritchken, Brian and Jackson, Brendon and Hu, Kelvin and Pancholi, Meghna and He, Yuan and Clancy, Brett and Colen, Chris and Wen, Fukang and Leung, Catherine and Wang, Siyuan and Zaruvinsky, Leon and Espinosa, Mateo and Lin, Rick and Liu, Zhongling and Padilla, Jake and Delimitrou, Christina},
title = {{An Open-Source Benchmark Suite for Microservices and Their Hardware-Software Implications for Cloud \& Edge Systems}},
booktitle = {Proceedings of the Twenty-Fourth International Conference on Architectural Support for Programming Languages and Operating Systems (ASPLOS'19)},
year = {2019}
}

@inproceedings{dcperf,
author = {Su, Wei and Dhanotia, Abhishek and Torres, Carlos and Gandhi, Jayneel and Gholkar, Neha and Kanaujia, Shobhit and Naumov, Maxim and Subramanian, Kalyan and Andrei, Valentin and Yuan, Yifan and Tang, Chunqiang},
title = {{DCPerf: An Open-Source, Battle-Tested Performance Benchmark Suite for Datacenter Workloads}},
year = {2025},
booktitle = {Proceedings of the 52nd Annual International Symposium on Computer Architecture (ISCA'25)}
}

@inproceedings{alibabaTraces,
  title={{Characterizing Microservice Dependency and Performance: Alibaba Trace Analysis}},
  author={Luo, Shutian and Xu, Huanle and Lu, Chengzhi and Ye, Kejiang and Xu, Guoyao and Zhang, Liping and Ding, Yu and He, Jian and Xu, Chengzhong},
  booktitle={Proceedings of the ACM Symposium on Cloud Computing (SoCC'21)},
  year={2021}
}

@inproceedings {metaMicro,
author = {Darby Huye and Yuri Shkuro and Raja R. Sambasivan},
title = {{Lifting the veil on {Meta{\textquoteright}s} microservice architecture: Analyses of topology and request workflows}},
booktitle = {Proceedings of the USENIX Annual Technical Conference (USENIX ATC'23)},
year = {2023},
}

@misc{netflix,
author = {Ketan Varshneya},
title = {{Understanding design of microservices architecture at Netflix}},
year = {2021},
howpublished = {\url{https://www.techaheadcorp.com/blog/design-of-microservices-architecture-at-netflix/}}
}

@inproceedings{luo2022Prediction,
  title={{The Power of Prediction: Microservice Auto Scaling via Workload Learning}},
  author={Luo, Shutian and Xu, Huanle and Ye, Kejiang and Xu, Guoyao and Zhang, Liping and Yang, Guodong and Xu, Chengzhong},
  booktitle={Proceedings of the ACM Symposium on Cloud Computing (SoCC'22)},
  year={2022}
}

@article{scaling,
title = {{Scaling equations for the accurate prediction of CMOS device performance from 180nm to 7nm}},
journal = {{Integration the VLSI journal}},
year = {2017},
author = {Aaron Stillmaker and Bevan Baas},
}

@inproceedings{warehouseScaleComp,
author = {Kanev, Svilen and Darago, Juan Pablo and Hazelwood, Kim and Ranganathan, Parthasarathy and Moseley, Tipp and Wei, Gu-Yeon and Brooks, David},
title = {{Profiling a warehouse-scale computer}},
year = {2015},
booktitle = {Proceedings of the 42nd Annual International Symposium on Computer Architecture (ISCA'15)},
}

@inproceedings{accelerometer,
author = {Sriraman, Akshitha and Dhanotia, Abhishek},
title = {{Accelerometer: Understanding Acceleration Opportunities for Data Center Overheads at Hyperscale}},
year = {2020},
booktitle = {Proceedings of the Twenty-Fifth International Conference on Architectural Support for Programming Languages and Operating Systems (ASPLOS'20)},
}

@misc{mahar2023workloadbehaviordrivenmemory,
      title={{Workload Behavior Driven Memory Subsystem Design for Hyperscale}}, 
      author={Suyash Mahar and Hao Wang and Wei Shu and Abhishek Dhanotia},
      year={2023},
      eprint={2303.08396},
      archivePrefix={arXiv},
      primaryClass={cs.DC},
      url={https://arxiv.org/abs/2303.08396}, 
}

@inproceedings{smartoclock,
    author = {Stojkovic, Jovan and Misra, Pulkit and Goiri, Inigo and Whitlock, Sam and Choukse, Esha and Das, Mayukh and Bansal, Chetan and Lee, Jason and Sun, Zoey and Qiu, Haoran and Zimmermann, Reed and Samal, Savyasachi and Warrier, Brijesh and Raniwala, Ashish and Bianchini, Ricardo},
    title = {{SmartOClock: Workload- and Risk-Aware Overclocking in the Cloud}},
    year = {2024},
    booktitle = {Proceedings of the 51st Annual International Symposium on Computer Architecture (ISCA '24)}
}

@inproceedings{tapas,
  title={{TAPAS: Thermal-and power-aware scheduling for LLM inference in cloud platforms}},
  author={Stojkovic, Jovan and Zhang, Chaojie and Goiri, {\'I}{\~n}igo and Choukse, Esha and Qiu, Haoran and Fonseca, Rodrigo and Torrellas, Josep and Bianchini, Ricardo},
  booktitle={Proceedings of the 30th ACM International Conference on Architectural Support for Programming Languages and Operating Systems (ASPLOS'25)},
  year={2025}
}

@INPROCEEDINGS{googlewebsearch,
  author={Ayers, Grant and Ahn, Jung Ho and Kozyrakis, Christos and Ranganathan, Parthasarathy},
  booktitle={Proceedings of the IEEE International Symposium on High Performance Computer Architecture (HPCA'18)},
  title={{Memory Hierarchy for Web Search}},
  year={2018}
}

@inproceedings{dynamoFB,
  author={Wu, Qiang and Deng, Qingyuan and Ganesh, Lakshmi and Hsu, Chang-Hong and Jin, Yun and Kumar, Sanjeev and Li, Bin and Meza, Justin and Song, Yee Jiun},
  booktitle={Proceedings of the 43rd Annual International Symposium on Computer Architecture (ISCA '16)},
  title={{Dynamo: Facebook's Data Center-Wide Power Management System}},
  year={2016}
}

@misc{k8s,
title = {{ Production-Grade Container Orchestration}},
author = {{Kubernetes}},
howpublished = {{https://kubernetes.io/}},
year = {2026}
}

@misc{azureTracesV2,
    author = {{Microsoft Azure}},
    title = "{Azure Public Dataset Version 2: VM Trace}",
    howpublished = {
\url{https://github.com/Azure/AzurePublicDataset/blob/master/AzurePublicDatasetV2.md}}
}

@inproceedings{dynamollm,
author = {Stojkovic, Jovan and Choukse, Esha and Zhang, Chaojie and Goiri, Inigo and Torrellas, Josep},
title = {{DynamoLLM: Designing LLM Inference Clusters for
Performance and Energy Efficiency}},
year = {2025},
booktitle = {Proceedings of the IEEE International Symposium on High-Performance Computer Architecture (HPCA '25)}
}

@inproceedings{thunderbolt,
    title = {{Thunderbolt: Throughput-Optimized, Quality-of-Service-Aware Power Capping at Scale}},
    author = {Shaohong Li and Xi Wang and Xiao Zhang and Vasileios Kontorinis and Sreekumar Kodakara and David Lo and Parthasarathy Ranganathan},
    booktitle = {Proceedings of the 14th USENIX Symposium on Operating Systems Design and Implementation (OSDI '20)},
    year = {2020},
}

@inproceedings{sinanML,
author = {Zhang, Yanqi and Hua, Weizhe and Zhou, Zhuangzhuang and Suh, G. Edward and Delimitrou, Christina},
title = {{Sinan: ML-Based and QoS-Aware Resource Management for Cloud Microservices}},
year = {2021},
booktitle = {Proceedings of the 26th International Conference on Architectural Support for Programming Languages and Operating Systems (ASPLOS '21)}
}

@inproceedings {protean,
author = {Ori Hadary and Luke Marshall and Ishai Menache and Abhisek Pan and Esaias E Greeff and David Dion and Star Dorminey and Shailesh Joshi and Yang Chen and Mark Russinovich and Thomas Moscibroda},
title = {{Protean: {VM} Allocation Service at Scale}},
booktitle = {Proceedings of the 14th USENIX Symposium on Operating Systems Design and Implementation (OSDI 20)},
year = {2020},
}

@inproceedings{resourcecentral,
author = {Cortez, Eli and Bonde, Anand and Muzio, Alexandre and Russinovich, Mark and Fontoura, Marcus and Bianchini, Ricardo},
title = {{Resource Central: Understanding and Predicting Workloads for Improved Resource Management in Large Cloud Platforms}},
year = {2017},
booktitle = {Proceedings of the 26th Symposium on Operating Systems Principles},
series = {SOSP '17}
}

@misc{schick2023toolformerlanguagemodelsteach,
      title={{Toolformer: Language Models Can Teach Themselves to Use Tools}},
      author={Timo Schick and Jane Dwivedi-Yu and Roberto Dessì and Roberta Raileanu and Maria Lomeli and Luke Zettlemoyer and Nicola Cancedda and Thomas Scialom},
      year={2023},
      eprint={2302.04761},
      archivePrefix={arXiv},
      primaryClass={cs.CL},
      url={https://arxiv.org/abs/2302.04761},
}

@misc{novikov2025alphaevolvecodingagentscientific,
      title={{AlphaEvolve: A coding agent for scientific and algorithmic discovery}},
      author={Alexander Novikov and Ngân Vũ and Marvin Eisenberger and Emilien Dupont and Po-Sen Huang and Adam Zsolt Wagner and Sergey Shirobokov and Borislav Kozlovskii and Francisco J. R. Ruiz and Abbas Mehrabian and M. Pawan Kumar and Abigail See and Swarat Chaudhuri and George Holland and Alex Davies and Sebastian Nowozin and Pushmeet Kohli and Matej Balog},
      year={2025},
      eprint={2506.13131},
      archivePrefix={arXiv},
      primaryClass={cs.AI},
      url={https://arxiv.org/abs/2506.13131},
}

@misc{liu2023codegeneratedchatgptreally,
      title={{Is Your Code Generated by ChatGPT Really Correct? Rigorous Evaluation of Large Language Models for Code Generation}},
      author={Jiawei Liu and Chunqiu Steven Xia and Yuyao Wang and Lingming Zhang},
      year={2023},
      eprint={2305.01210},
      archivePrefix={arXiv},
      primaryClass={cs.SE},
      url={https://arxiv.org/abs/2305.01210},
}

@article{mirhoseini2021graph,
  author    = "Azalia Mirhoseini and Anna Goldie and Mustafa Yazgan and Joe Wenjie Jiang and Ebrahim Songhori and Shen Wang and Young-Joon Lee and Eric Johnson and Omkar Pathak and Azade Nazi and Jiwoo Pak and Andy Tong and Kavya Srinivasa and William Hang and Emre Tuncer and Quoc V. Le and James Laudon and Richard Ho and Roger Carpenter and Jeff Dean",
  title     = "A Graph Placement Methodology for Fast Chip Design",
  journal   = "Nature",
  year      = "2021"
}

@inproceedings{prefetcher,
  author    = "Milad Hashemi and Kevin Swersky and Jamie A. Smith and Grant Ayers and Heiner Litz and Jichuan Chang and Christos Kozyrakis and Parthasarathy Ranganathan",
  title     = "Learning Memory Access Patterns",
  booktitle = "Proceedings of the 35th International Conference on Machine Learning (ICML)",
  year      = "2018"
}

@inproceedings{distributedcaching,
author = {Mao, Ziming and Ellithorpe, Jonathan and Adya, Atul and Iyer, Rishabh and Zaharia, Matei and Shenker, Scott and Stoica, Ion},
title = {{Rethinking the Cost of Distributed Caches for Datacenter Services}},
year = {2025},
booktitle = {Proceedings of the 24th ACM Workshop on Hot Topics in Networks},
series = {HotNets '25}
}

@misc{cemri2026adaevolveadaptivellmdriven,
      title={{AdaEvolve: Adaptive LLM Driven Zeroth-Order Optimization}},
      author={Mert Cemri and Shubham Agrawal and Akshat Gupta and Shu Liu and Audrey Cheng and Qiuyang Mang and Ashwin Naren and Lutfi Eren Erdogan and Koushik Sen and Matei Zaharia and Alex Dimakis and Ion Stoica},
      year={2026},
      eprint={2602.20133},
      archivePrefix={arXiv},
      primaryClass={cs.NE},
      url={https://arxiv.org/abs/2602.20133},
}

@inproceedings{autopilot,
author = {Rzadca, Krzysztof and Findeisen, Pawel and Swiderski, Jacek and Zych, Przemyslaw and Broniek, Przemyslaw and Kusmierek, Jarek and Nowak, Pawel and Strack, Beata and Witusowski, Piotr and Hand, Steven and Wilkes, John},
title = {{Autopilot: workload autoscaling at Google}},
year = {2020},
booktitle = {Proceedings of the Fifteenth European Conference on Computer Systems},
series = {EuroSys '20}
}

@inproceedings{perceptron,
author = {Jim\'{e}nez, Daniel A. and Lin, Calvin},
title = {{Dynamic Branch Prediction with Perceptrons}},
year = {2001},
booktitle = {Proceedings of the 7th International Symposium on High-Performance Computer Architecture},
pages = {197},
series = {HPCA '01}
}

@InProceedings{pmlr-v80-hashemi18a,
  title = 	 {{Learning Memory Access Patterns}},
  author =       {Hashemi, Milad and Swersky, Kevin and Smith, Jamie and Ayers, Grant and Litz, Heiner and Chang, Jichuan and Kozyrakis, Christos and Ranganathan, Parthasarathy},
  booktitle = 	 {Proceedings of the 35th International Conference on Machine Learning},
  year = 	 {2018},
}

@inproceedings{rl-device,
author = {Mirhoseini, Azalia and Pham, Hieu and Le, Quoc V. and Steiner, Benoit and Larsen, Rasmus and Zhou, Yuefeng and Kumar, Naveen and Norouzi, Mohammad and Bengio, Samy and Dean, Jeff},
title = {{Device placement optimization with reinforcement learning}},
year = {2017},
booktitle = {Proceedings of the 34th International Conference on Machine Learning - Volume 70},
series = {ICML'17}
}

@inproceedings{replacement,
author = {Shi, Zhan and Huang, Xiangru and Jain, Akanksha and Lin, Calvin},
title = {{Applying Deep Learning to the Cache Replacement Problem}},
year = {2019},
booktitle = {Proceedings of the 52nd Annual IEEE/ACM International Symposium on Microarchitecture},
series = {MICRO-52}
}

@article{gptuner,
author = {Lao, Jiale and Wang, Yibo and Li, Yufei and Wang, Jianping and Zhang, Yunjia and Cheng, Zhiyuan and Chen, Wanghu and Tang, Mingjie and Wang, Jianguo},
title = {{GPTuner: A Manual-Reading Database Tuning System via GPT-Guided Bayesian Optimization}},
year = {2024},
issue_date = {April 2024},
publisher = {VLDB Endowment},
volume = {17},
number = {8},
journal = {Proc. VLDB Endow.},
month = apr,
pages = {1939–1952},
numpages = {14}
}

@misc{chen2021evaluatinglargelanguagemodels,
      title={{Evaluating Large Language Models Trained on Code}},
      author={Mark Chen and Jerry Tworek and Heewoo Jun and Qiming Yuan and Henrique Ponde de Oliveira Pinto and Jared Kaplan and Harri Edwards and Yuri Burda and Nicholas Joseph and Greg Brockman and Alex Ray and Raul Puri and Gretchen Krueger and Michael Petrov and Heidy Khlaaf and Girish Sastry and Pamela Mishkin and Brooke Chan and Scott Gray and Nick Ryder and Mikhail Pavlov and Alethea Power and Lukasz Kaiser and Mohammad Bavarian and Clemens Winter and Philippe Tillet and Felipe Petroski Such and Dave Cummings and Matthias Plappert and Fotios Chantzis and Elizabeth Barnes and Ariel Herbert-Voss and William Hebgen Guss and Alex Nichol and Alex Paino and Nikolas Tezak and Jie Tang and Igor Babuschkin and Suchir Balaji and Shantanu Jain and William Saunders and Christopher Hesse and Andrew N. Carr and Jan Leike and Josh Achiam and Vedant Misra and Evan Morikawa and Alec Radford and Matthew Knight and Miles Brundage and Mira Murati and Katie Mayer and Peter Welinder and Bob McGrew and Dario Amodei and Sam McCandlish and Ilya Sutskever and Wojciech Zaremba},
      year={2021},
      eprint={2107.03374},
      archivePrefix={arXiv},
      primaryClass={cs.LG},
      url={https://arxiv.org/abs/2107.03374},
}

@misc{deepseek-coder,
      title={{DeepSeek-Coder: When the Large Language Model Meets Programming -- The Rise of Code Intelligence}},
      author={Daya Guo and Qihao Zhu and Dejian Yang and Zhenda Xie and Kai Dong and Wentao Zhang and Guanting Chen and Xiao Bi and Y. Wu and Y. K. Li and Fuli Luo and Yingfei Xiong and Wenfeng Liang},
      year={2024},
      eprint={2401.14196},
      archivePrefix={arXiv},
      primaryClass={cs.SE},
      url={https://arxiv.org/abs/2401.14196},
}

@misc{cummins2024,
      title={{Large Language Models for Compiler Optimization}},
      author={Chris Cummins and Volker Seeker and Dejan Grubisic and Mostafa Elhoushi and Youwei Liang and Baptiste Roziere and Jonas Gehring and Fabian Gloeckle and Kim Hazelwood and Gabriel Synnaeve and Hugh Leather},
      year={2023},
      eprint={2309.07062},
      archivePrefix={arXiv},
      primaryClass={cs.PL},
      url={https://arxiv.org/abs/2309.07062},
}

@inproceedings{borg,
author = {Verma, Abhishek and Pedrosa, Luis and Korupolu, Madhukar and Oppenheimer, David and Tune, Eric and Wilkes, John},
title = {{Large-scale cluster management at Google with Borg}},
year = {2015},
booktitle = {Proceedings of the Tenth European Conference on Computer Systems},
articleno = {18},
numpages = {17},
location = {Bordeaux, France},
series = {EuroSys '15}
}

@misc{react,
      title={{ReAct: Synergizing Reasoning and Acting in Language Models}},
      author={Shunyu Yao and Jeffrey Zhao and Dian Yu and Nan Du and Izhak Shafran and Karthik Narasimhan and Yuan Cao},
      year={2023},
      eprint={2210.03629},
      archivePrefix={arXiv},
      primaryClass={cs.CL},
      url={https://arxiv.org/abs/2210.03629},
}

@misc{agenticAI,
  title={{Practices for Governing Agentic AI Systems}},
  author={Yonadav Shavit and Sandhini Agarwal and Miles Brundage and Steven O'Keefe and Rosie Campbell and Teddy Lee and Pamela Mishkin and Tyna Eloundou and Alan Hickey and Katarina Slama and Lama Ahmad and Paul McMillan and Alex Beutel and Alexandre Passos and David G. Robinson},
  howpublished ={\url{https://api.semanticscholar.org/CorpusID:266312974}}
}

@misc{starcoder,
      title={{StarCoder: may the source be with you!}},
      author={Raymond Li and Loubna Ben Allal and Yangtian Zi and Niklas Muennighoff and Denis Kocetkov and Chenghao Mou and Marc Marone and Christopher Akiki and Jia Li and Jenny Chim and Qian Liu and Evgenii Zheltonozhskii and Terry Yue Zhuo and Thomas Wang and Olivier Dehaene and Mishig Davaadorj and Joel Lamy-Poirier and Jo\~{a}o Monteiro and Oleh Shliazhko and Nicolas Gontier and Nicholas Meade and Armel Zebaze and Ming-Ho Yee and Logesh Kumar Umapathi and Jian Zhu and Benjamin Lipkin and Muhtasham Oblokulov and Zhiruo Wang and Rudra Murthy and Jason Stillerman and Siva Sankalp Patel and Dmitry Abulkhanov and Marco Zocca and Manan Dey and Zhihan Zhang and Nour Fahmy and Urvashi Bhattacharyya and Wenhao Yu and Swayam Singh and Sasha Luccioni and Paulo Villegas and Maxim Kunakov and Fedor Zhdanov and Manuel Romero and Tony Lee and Nadav Timor and Jennifer Ding and Claire Schlesinger and Hailey Schoelkopf and Jan Ebert and Tri Dao and Mayank Mishra and Alex Gu and Jennifer Robinson and Carolyn Jane Anderson and Brendan Dolan-Gavitt and Danish Contractor and Siva Reddy and Daniel Fried and Dzmitry Bahdanau and Yacine Jernite and Carlos Mu\~{n}oz Ferrandis and Sean Hughes and Thomas Wolf and Arjun Guha and Leandro von Werra and Harm de Vries},
      year={2023},
      eprint={2305.06161},
      archivePrefix={arXiv},
      primaryClass={cs.CL},
      url={https://arxiv.org/abs/2305.06161},
}

@inproceedings{xia2023apr,
author = {Xia, Chunqiu Steven and Wei, Yuxiang and Zhang, Lingming},
title = {{Automated Program Repair in the Era of Large Pre-Trained Language Models}},
year = {2023},
isbn = {9781665457019},
publisher = {IEEE Press},
url = {https://doi.org/10.1109/ICSE48619.2023.00129},
doi = {10.1109/ICSE48619.2023.00129},
booktitle = {{Proceedings of the 45th International Conference on Software Engineering}},
pages = {1482--1494},
numpages = {13},
location = {Melbourne, Victoria, Australia},
series = {ICSE '23}
}

@article{Levin_2025,
   title={{ChatDBG: Augmenting Debugging with Large Language Models}},
   volume={2},
   ISSN={2994-970X},
   url={http://dx.doi.org/10.1145/3729355},
   DOI={10.1145/3729355},
   number={FSE},
   journal={Proceedings of the ACM on Software Engineering},
   publisher={Association for Computing Machinery (ACM)},
   author={Levin, Kyla H. and van Kempen, Nicolas and Berger, Emery D. and Freund, Stephen N.},
   year={2025},
   month=jun, pages={1892--1913} }

@misc{thakur2023verigen,
      title={{VeriGen: A Large Language Model for Verilog Code Generation}},
      author={Shailja Thakur and Baleegh Ahmad and Hammond Pearce and Benjamin Tan and Brendan Dolan-Gavitt and Ramesh Karri and Siddharth Garg},
      year={2023},
      eprint={2308.00708},
      archivePrefix={arXiv},
      primaryClass={cs.PL},
      url={https://arxiv.org/abs/2308.00708},
}

@misc{sweagent,
      title={{SWE-agent: Agent-Computer Interfaces Enable Automated Software Engineering}},
      author={John Yang and Carlos E. Jimenez and Alexander Wettig and Kilian Lieret and Shunyu Yao and Karthik Narasimhan and Ofir Press},
      year={2024},
      eprint={2405.15793},
      archivePrefix={arXiv},
      primaryClass={cs.SE},
      url={https://arxiv.org/abs/2405.15793},
}

@misc{metagpt,
      title={{MetaGPT: Meta Programming for A Multi-Agent Collaborative Framework}},
      author={Sirui Hong and Mingchen Zhuge and Jiaqi Chen and Xiawu Zheng and Yuheng Cheng and Ceyao Zhang and Jinlin Wang and Zili Wang and Steven Ka Shing Yau and Zijuan Lin and Liyang Zhou and Chenyu Ran and Lingfeng Xiao and Chenglin Wu and J\"{u}rgen Schmidhuber},
      year={2024},
      eprint={2308.00352},
      archivePrefix={arXiv},
      primaryClass={cs.AI},
      url={https://arxiv.org/abs/2308.00352},
}

@article{llmcodegen,
author = {Jiang, Juyong and Wang, Fan and Shen, Jiasi and Kim, Sungju and Kim, Sunghun},
title = {{A Survey on Large Language Models for Code Generation}},
year = {2026},
journal = {ACM Trans. Softw. Eng. Methodol.},
month = jan,
articleno = {58},
numpages = {72}
}

@misc{lin2025ecollmdrivenefficientcode,
      title={{ECO: An LLM-Driven Efficient Code Optimizer for Warehouse Scale Computers}},
      author={Hannah Lin and Martin Maas and Maximilian Roquemore and Arman Hasanzadeh and Fred Lewis and Yusuf Simonson and Tzu-Wei Yang and Amir Yazdanbakhsh and Deniz Altinb\"{u}ken and Florin Papa and Maggie Nolan Edmonds and Aditya Patil and Don Schwarz and Satish Chandra and Chris Kennelly and Milad Hashemi and Parthasarathy Ranganathan},
      year={2025},
      eprint={2503.15669},
      archivePrefix={arXiv},
      primaryClass={cs.SE},
      url={https://arxiv.org/abs/2503.15669},
}

@misc{kernelevolve,
      title={{KernelEvolve: Scaling Agentic Kernel Coding for Heterogeneous AI Accelerators at Meta}},
      author={Gang Liao and Hongsen Qin and Ying Wang and Alicia Golden and Michael Kuchnik and Yavuz Yetim and Jia Jiunn Ang and Chunli Fu and Yihan He and Samuel Hsia and Zewei Jiang and Dianshi Li and Uladzimir Pashkevich and Varna Puvvada and Feng Shi and Matt Steiner and Ruichao Xiao and Nathan Yan and Xiayu Yu and Zhou Fang and Roman Levenstein and Kunming Ho and Haishan Zhu and Alec Hammond and Richard Li and Ajit Mathews and Kaustubh Gondkar and Abdul Zainul-Abedin and Ketan Singh and Hongtao Yu and Wenyuan Chi and Barney Huang and Sean Zhang and Noah Weller and Zach Marine and Wyatt Cook and Carole-Jean Wu and Gaoxiang Liu},
      year={2026},
      eprint={2512.23236},
      archivePrefix={arXiv},
      primaryClass={cs.LG},
      url={https://arxiv.org/abs/2512.23236},
}

@misc{aicudaengineer,
      title={{Towards Robust Agentic CUDA Kernel Benchmarking, Verification, and Optimization}},
      author={Robert Tjarko Lange and Qi Sun and Aaditya Prasad and Maxence Faldor and Yujin Tang and David Ha},
      year={2025},
      eprint={2509.14279},
      archivePrefix={arXiv},
      primaryClass={cs.SE},
      url={https://arxiv.org/abs/2509.14279},
}

@misc{geak,
      title={{Geak: Introducing Triton Kernel AI Agent \& Evaluation Benchmarks}},
      author={Jianghui Wang and Vinay Joshi and Saptarshi Majumder and Xu Chao and Bin Ding and Ziqiong Liu and Pratik Prabhanjan Brahma and Dong Li and Zicheng Liu and Emad Barsoum},
      year={2025},
      eprint={2507.23194},
      archivePrefix={arXiv},
      primaryClass={cs.CL},
      url={https://arxiv.org/abs/2507.23194},
}

@misc{kevin,
  title={{Kevin: Multi-Turn Reinforcement Learning for Writing CUDA Kernels}},
  author={Baronio, Carlo and Marsella, Pietro and Pan, Ben and others},
  year={2025},
  howpublished={arXiv preprint arXiv:2507.11948},
  note={\url{https://arxiv.org/abs/2507.11948}}
}

@inproceedings{skydiscover,
author = {Liu, Shu and Cemri, Mert and Agarwal, Shubham and Krentsel, Alexander and Naren, Ashwin and Mang, Qiuyang and Li, Zhifei and Gupta, Akshat and Maheswaran, Monishwaran and Cheng, Audrey and Pan, Melissa and Boneh, Ethan and Ramchandran, Kannan and Sen, Koushik and Zaharia, Matei and Dimakis, Alexandros G. and Stoica, Ion},
title = {{SkyDiscover: A Flexible, Adaptive Framework for AI-Driven Scientific and Algorithmic Discovery}},
year = {2026},
isbn = {9798400724152},
publisher = {Association for Computing Machinery},
address = {New York, NY, USA},
url = {https://doi.org/10.1145/3786335.3813221},
doi = {10.1145/3786335.3813221},
booktitle = {Proceedings of the ACM Conference on AI and Agentic Systems},
pages = {1223--1227},
numpages = {5},
series = {CAIS '26}
}

@misc{microsoftchip,
    title = {{Azure Cobalt processor-based Virtual Machines}},
    author = {{Microsoft Azure}},
    year = {2026},
    howpublished = {\url{https://learn.microsoft.com/en-us/azure/virtual-machines/sizes/cobalt-overview}}
}

@misc{googlechip,
    title = {{Introducing Google Axion Processors, our new Arm-based CPUs}},
    author = {{Google}},
    year = {2026},
    howpublished = {\url{https://cloud.google.com/blog/products/compute/introducing-googles-new-arm-based-cpu}}
}

@misc{amazonchip,
    title = {{AWS Graviton Processor}},
    year = {2026},
    author = {{Amazon AWS}},
    howpublished = {\url{https://aws.amazon.com/pm/ec2-graviton/?trk=68d7c491-4ff9-4549-b7e7-778fc8ff3952}}
}

@inproceedings{microsoftworkloads,
author = {Parayil, Anjaly and Zhang, Jue and Qin, Xiaoting and Goiri, \'{I}\~{n}igo and Huang, Lexiang and Zhu, Timothy and Bansal, Chetan},
title = {{Towards Workload-aware Cloud Efficiency: A Large-scale Empirical Study of Cloud Workload Characteristics}},
year = {2025},
booktitle = {Proceedings of the 16th ACM/SPEC International Conference on Performance Engineering (ICPE '25)},
}

\end{document}